\documentclass[]{article}

\usepackage{amsmath,amsfonts}
\usepackage{algorithm}
\usepackage{array}
\usepackage[caption=false,font=normalsize,labelfont=sf,textfont=sf]{subfig}
\usepackage{textcomp}
\usepackage{stfloats}
\usepackage{url}
\usepackage{verbatim}
\usepackage{graphicx}
\usepackage{cite}

\usepackage[utf8]{inputenc}
\usepackage[T1]{fontenc}
\usepackage{lmodern}
\usepackage{microtype}

\usepackage{amsmath,amssymb,amsfonts}
\usepackage{amsthm}
\usepackage{mathrsfs}
\usepackage{physics}

\usepackage{amsthm}

\usepackage{xcolor}

\usepackage{graphicx}

\usepackage{tikz}
\usetikzlibrary{
	positioning,
	arrows.meta,
	backgrounds,
	calc,
	decorations.pathreplacing,
	shadows.blur,
	shadows
}

\usepackage[margin=2cm]{geometry}
\usepackage{booktabs}   
\usepackage{multirow}
\usepackage{makecell}
\usepackage{tabularx}
\usepackage{longtable}
\usepackage{lscape}
\usepackage{rotating}

\usepackage{algorithm}
\usepackage{algorithmicx}
\usepackage{algpseudocode}

\usepackage{listings}
\usepackage{textcomp}   

\usepackage{manyfoot}

\tikzset{
	lzcbox/.style={
		rectangle, draw=black!80, thick,
		rounded corners=3pt,
		minimum width=2.8cm, minimum height=1cm,
		font=\scriptsize, align=center,
		fill=blue!5, drop shadow
	}
}

\usepackage{authblk}

\title{Learning Internal Biological Neuron Parameters and Complexity-Based Encoding for Improved Spiking Neural Networks Performance}

\author{Zofia Rudnicka}
\author{Janusz Szczepanski}
\author{Agnieszka Pregowska}
\affil{Institute of Fundamental Technological Research, Polish Academy of Sciences,	Pawinskiego~5B, 02--106 Warsaw, Poland}

\date{} 

\begin{document}
\maketitle
	\begin{abstract}
	This study introduces a novel learning paradigm for spiking neural networks (SNNs) by replacing the traditional perceptron-inspired neuron abstraction with biologically motivated neuron models in which not only synaptic weights but also intrinsic neuronal parameters are jointly optimized during training. We evaluate two SNN architectures, one based on standard leaky integrate-and-fire (LIF) neurons and another employing a meta-neuron mode, and compare settings with fixed versus learnable intrinsic parameters. As a second contribution, we propose a biologically inspired classification framework that integrates SNN dynamics with Lempel-Ziv complexity (LZC), a measure closely related to entropy rate, enabling efficient and interpretable classification of spatiotemporal spike data.
	
	We consider multiple learning algorithms, including backpropagation with surrogate gradients, spike-timing-dependent plasticity (STDP), and the Tempotron rule, and train on spike trains generated from Poisson processes, which are widely adopted in computational neuroscience as a standard stochastic model for neuronal spike generation due to their analytical tractability and empirical relevance to irregular cortical firing patterns. Experiments show that learning intrinsic neuron parameters in addition to synaptic weights improves classification accuracy by up to 13.50 percentage points for LIF-based SNNs and up to 8.50 percentage points for meta-neuron models, relative to baselines where only network size and learning rate are tuned. Moreover, the proposed SNN-LZC classifier reaches accuracies up to 99.50\% while maintaining sub-millisecond inference latency and competitive energy consumption.
	
	In addition to empirical evaluation, we provide a quantitative theoretical justification for the observed gains by formalizing the effect of enlarging the hypothesis class through learnable intrinsic dynamics and deriving descent guarantees for intrinsic-parameter updates under standard smoothness (Lipschitz-gradient) assumptions. This analysis links neuron-intrinsic optimization to provable improvements in the surrogate objective and supports the practical relevance of jointly optimizing internal neuron dynamics and complexity-based readout.
\end{abstract}

\section*{keywords}
	Spiking Neural Networks, Learning Algorithms, Neuron Models, Lempel-Ziv Complexity, Spike Trains, Classification.

\section{Introduction}
\par
Spiking Neural Networks (SNNs) have emerged as promising candidates to address the limitations of conventional artificial neural networks (ANNs), particularly in real-world applications where temporal precision and energy efficiency are critical \cite{Tang2025}. Unlike traditional ANNs, SNNs more closely approximate the behavior of biological neurons by transmitting information through discrete spike events, thus capturing both the spatial and temporal dynamics of neural activity \cite{Stan2024}. This biologically grounded architecture positions SNNs as potential computational models capable of approaching the efficiency and adaptability of the human brain \cite{Gao2025}.
\par
Despite their promise, training SNNs remains a significant challenge due to their non-linear, event-driven dynamics and the non-differentiable nature of spike generation \cite{Yi2023,Hu2025}. Traditional gradient-based methods such as backpropagation, which iteratively minimize a loss function by computing gradients of continuous activation values, are not well-suited to the discrete and temporally sparse nature of spiking activity \cite{Wu2018}. Moreover, these conventional arithmetic-based learning rules are often biologically implausible and inefficient when applied to spiking architectures \cite{bassler2022,Ding2025}. Instead, the optimization of intrinsic neuron parameters and the adoption of biologically inspired temporal coding schemes play a critical role in improving the performance and interpretability of SNNs \cite{Saglam2023}. Temporal coding, in particular, has been shown to transmit information more efficiently than rate coding, potentially reducing the number of spikes required for accurate representation and decision-making \cite{Rudnicka2024,Jin2025}.
\par
On the other hand, concepts from Information Theory (IT), like Lempel-Ziv Complexty (LZC) provide powerful tools for analyzing neural data and understanding the computational principles underlying brain-like processing \cite{Shannon1948,LempelZiv1976,Bossomaier2016}. LZC offers a multidimensional perspective on neural signals \cite{Arnold2013,Pregowska2019}. This quantity not only account for the probabilistic characteristics of spiking data but also capture the underlying structural patterns and temporal regularities. This dual sensitivity allows them to quantify the degree of randomness in neural activity while simultaneously identifying deterministic features, i.e. an essential property for characterizing the outputs of SNNs.
\par
In this study, we propose a novel modification to the spiking neural network by replacing the traditional perceptron neuron model with a biologically inspired meta neuron, in which internal parameters are jointly learned during training. We hypothesize and demonstrate that this biologically motivated enhancement can significantly improve SNN classification accuracy. To validate this, we implement two SNN architectures: one based on standard leaky integrate-and-fire (LIF) neurons \cite{Dutta2017} and the other on the proposed meta neuron model \cite{Cheng2023,Bansal2024}. As a second key contribution, we introduce a novel biologically inspired classification framework that integrates SNNs with Lempel-Ziv complexity. By leveraging the temporal precision and biological plausibility of SNNs together with LZC’s ability to quantify structural complexity, this framework provides efficient and interpretable analysis of spatiotemporal neural patterns for classification tasks, namely a capability not present in prior approaches. We further evaluate multiple learning algorithms, including backpropagation \cite{Wu2018}, spike-timing-dependent plasticity (STDP) \cite{Hao2020}, and the Tempotron learning rule \cite{Gutig2006}.

\begin{table*}[t]
	\centering
	\small
	\caption{Basic notation used in the study.}
	\label{tab:notation}
	\begin{tabular}{p{5.2cm} p{9.3cm}}
		\hline
		\textbf{Notation} & \textbf{Description} \\
		\hline
		$\mathbf{x} = [x_{1}, x_{2}, \ldots, x_{n}] \in \mathbb{R}^{n}$ 
		& Synaptic weight vector. \\
		
		$V_{m}(t) \in \mathbb{R}$ 
		& Membrane potential at time $t$. \\
		
		$\tau_{m} \in \mathbb{R}^{+}$ 
		& Membrane time constant. \\
		
		$I(t) \in \mathbb{R}$ 
		& Input current at time $t$. \\
		
		$t_{f} \in \mathbb{R}^{+}$ 
		& Neuron firing time. \\
		
		$S_{i}(t) \in \{0,1\}$ 
		& Binary-valued spike train at time $t$. \\
		
		$t_{i}^{k} \in \mathbb{R}^{+}$ 
		& Time of the $k$-th spike of neuron $i$. \\
		
		$\eta \in \mathbb{R}^{+}$ 
		& Learning rate used for weight updates. \\
		
		$\eta_{+}, \eta_{-} \in \mathbb{R}^{+}$ 
		& STDP time constants. \\
		
		$A_{+}, A_{-} \in \mathbb{R}^{+}$ 
		& Amplitudes of STDP weight changes. \\
		
		$t_{\mathrm{pre}}, t_{\mathrm{post}} \in \mathbb{R}^{+}$ 
		& Pre- and postsynaptic spike times. \\
		
		$K(t) : \mathbb{R} \rightarrow \mathbb{R}$ 
		& Kernel function mapping time to a real value. \\
		
		$t_{i}^{\mathrm{target}}, t_{i}^{\mathrm{actual}} \in \mathbb{R}^{+}$ 
		& Target and actual spike times of neuron $i$. \\
		
		$\mathcal{E} \in \mathbb{R}^{+}$ 
		& Error function value. \\
		
		$\frac{\partial \mathcal{E}}{\partial w_{i}} \in \mathbb{R}$ 
		& Gradient of the error with respect to weight $w_i$. \\
		\hline
	\end{tabular}
\end{table*}

\section{Notation}

An overview of the key notation is provided in Table \ref{tab:notation}.

\begin{figure*}[t]
	\centering
	\begin{tikzpicture}[
		block/.style={
			rectangle, rounded corners=4pt,
			minimum width=3.4cm, minimum height=1cm,
			font=\scriptsize\sffamily, align=center,
			draw=black, blur shadow
		},
		lifblock/.style={block, fill=cyan!20},
		metablock/.style={block, fill=blue!20},
		adaptblock/.style={block, fill=blue!10},
		arrow/.style={-{Latex[length=2mm]}, thick, draw=black!70},
		darrow/.style={-{Latex[length=2mm]}, thick, dashed, draw=blue!60},
		title/.style={font=\bfseries\small\sffamily, text=blue!80}
		]
		
		\node[title] at (0,6.2) {LIF Neuron};
		\node[title] at (7.2,6.2) {Meta-Neuron};
		
		\node[lifblock] (lif_in) at (0,5.4) {Input Current\\$I(t)$};
		\node[lifblock, below=0.6cm of lif_in] (lif_int)
		{Integration\\$\tau\,\dfrac{dV}{dt} = -V + R\,I(t)$};
		\node[lifblock, below=0.6cm of lif_int] (lif_th)
		{Threshold\\$V(t) \ge V_{\mathrm{th}}$};
		\node[lifblock, below=0.6cm of lif_th] (lif_out)
		{Output Spike\\$S(t) \in \{0,1\}$};
		
		\draw[arrow] (lif_in) -- (lif_int);
		\draw[arrow] (lif_int) -- (lif_th);
		\draw[arrow] (lif_th) -- (lif_out);
		
		\coordinate (below_out) at ($(lif_out.south) + (0,-0.4)$);
		\node[font=\tiny\sffamily\color{blue!60}, below=2pt of below_out]
		{$\circlearrowleft$ Reset: $V \leftarrow V_{\mathrm{reset}}$};
		\draw[darrow] (lif_out.south) to[out=-90, in=-90] (lif_int.south);
		
		\node[metablock] (meta_in) at (7.2,5.4) {Input Current\\$I(t)$};
		\node[metablock, below=0.6cm of meta_in] (meta_int)
		{Integration\\$\tau(t)\,\dfrac{dV}{dt} = -V + R\,I(t)$};
		\node[metablock, below=0.6cm of meta_int] (meta_th)
		{Threshold\\$V(t) \ge V_{\mathrm{th}}(t)$};
		\node[metablock, below=0.6cm of meta_th] (meta_out)
		{Output Spike\\$S(t) \in \{0,1\}$};
		
		\draw[arrow] (meta_in) -- (meta_int);
		\draw[arrow] (meta_int) -- (meta_th);
		\draw[arrow] (meta_th) -- (meta_out);
		
		\draw[darrow] (meta_out.east) to[out=0, in=-90] ++(1.1,1.1)
		to[out=90, in=0] (meta_int.east);
		\node[font=\tiny\sffamily\color{blue!60}, right=3pt of meta_out.east, anchor=west]
		{$\circlearrowleft$};
		
		\node[adaptblock, right=2.3cm of meta_int] (adapt_tau)
		{Adaptive Time Const.\\$\tau(t)$};
		\node[adaptblock, right=2.3cm of meta_th] (adapt_th)
		{Adaptive Threshold\\$V_{\mathrm{th}}(t)$};
		
		\draw[darrow] (adapt_tau.west) -- (meta_int.east);
		\draw[darrow] (adapt_th.west) -- (meta_th.east);
		
	\end{tikzpicture}
	
	\caption{Comparison of the standard Leaky Integrate-and-Fire (LIF) neuron model and the extended meta-neuron architecture. The LIF neuron integrates the input current $I(t)$ over time according to the leaky membrane equation $\tau \frac{dV}{dt} = -V + RI(t)$, where $\tau$ is the membrane time constant and $R$ is the membrane resistance. A spike is emitted when the membrane potential $V(t)$ exceeds a fixed threshold $V_{\mathrm{th}}$, followed by a reset to $V_{\mathrm{reset}}$. The meta-neuron preserves this core structure but incorporates dynamic, adaptive components. Specifically, the membrane time constant $\tau(t)$ and threshold $V_{\mathrm{th}}(t)$ can vary over time, enabling richer and more flexible temporal dynamics. More generally, the meta-neuron can be interpreted as computing a parameterized function $y=f(\mathbf{x};\boldsymbol{\theta})$, where the input vector $\mathbf{x}$ is mapped to an output $y$ through parameters $\boldsymbol{\theta}$, which may be adapted online through feedback mechanisms or learning algorithms.}
	\label{fig:lif_meta}
\end{figure*}
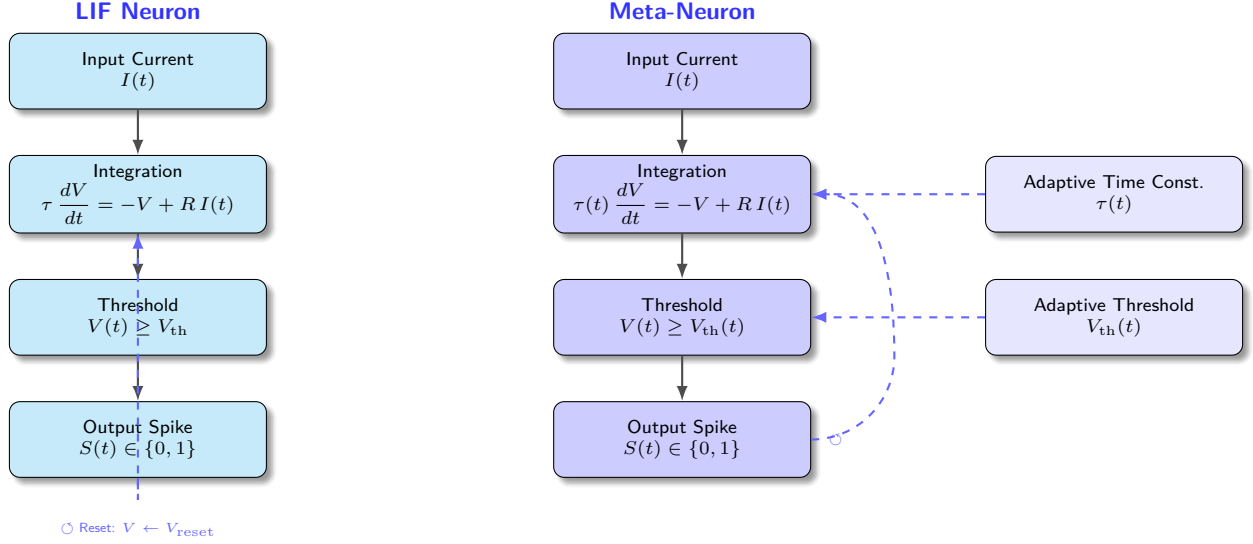

\section{Neuron Models}

The Leaky Integrate-and-Fire (LIF) neuron receives an input vector 
$\mathbf{x}(t) \in \mathbb{R}^n$, which is linearly combined with a weight vector $\mathbf{w} \in \mathbb{R}^n$ and a bias term $b \in \mathbb{R}$ to compute the total input current: \[ I(t) = \mathbf{w}^\top \mathbf{x}(t) + b. \] The membrane potential $V_m(t)$ evolves over time according to a leaky integration mechanism  governed by the membrane time constant $\tau_m$ and resistance $R_m$. When $V_m(t)$ exceeds a fixed threshold $V_{\mathrm{th}}$, a spike is emitted and the potential is reset to the resting potential $V_{\mathrm{reset}}$ \cite{Dutta2017,Yaya2025}.

In contrast, the meta-neuron extends this model by allowing key neuronal parameters to adapt over time \cite{Yao2024}. Specifically, both the membrane time constant $\tau_m(t)$ and the threshold $V_{\mathrm{th}}(t)$ are dynamically modulated based on the internal state of the network or external feedback. More abstractly, the metaneuron computes an output.
\[
y(t) = f\!\big(\mathbf{x}(t); \boldsymbol{\theta}(t)\big),
\]
where $\boldsymbol{\theta}(t)$ denotes a time-varying set of internal parameters subject to learning or contextual updates. This framework enables more flexible and expressive computations while preserving the spiking behavior characteristic of biologically inspired models.
\par
While the LIF model offers a biologically grounded mechanism for spike generation, its static parameters limit expressivity. To overcome this, we extend the LIF framework into a meta-neuron formulation, where parameters evolve dynamically and computations generalize beyond simple thresholding.  As illustrated in Figure~\ref{fig:lif_meta}, the meta-neuron preserves the core integration-threshold-reset structure of the LIF but augments it with time-varying parameters such as $\tau(t)$ and $V_{\text{th}}(t)$, or even fully parameterized update functions. This supports a broader class of computations, bridging spiking dynamics with principles from meta-learning and optimization.

\section{Spiking Neural Network}

\begin{figure*}
	\centering
	\begin{tikzpicture}[
		neuron/.style={circle, draw=black!70, thick, minimum size=1cm, inner sep=0pt, drop shadow, font=\scriptsize\bfseries, text=white},
		layerlabel/.style={font=\bfseries\small\sffamily},
		arr/.style={-{Latex[length=1.8mm]}, thick, draw=black!60},
		bit/.style={font=\scriptsize\ttfamily},
		lzcbox/.style={rectangle, draw=black!80, thick, rounded corners, minimum width=2.8cm, minimum height=1cm, font=\scriptsize, align=center, drop shadow}
		]
		
		\definecolor{blue1}{RGB}{173,216,230}  
		\definecolor{blue2}{RGB}{100,149,237}  
		\definecolor{blue3}{RGB}{70,130,180}   
		\definecolor{blue4}{RGB}{0,102,204}    
		\definecolor{blue5}{RGB}{0,76,153}     
		
		\def\xinput{0}
		\def\xhidden{3}
		\def\xoutput{6}
		\def\xlzc{9.3}
		\def\ydist{1.2}
		
		\node[layerlabel] at (\xinput,3.5) {Input Layer $X$};
		\node[layerlabel] at (\xhidden,3.5) {Hidden Layer $H$};
		\node[layerlabel] at (\xoutput,3.5) {Output Layer $Z$};
		
		\node[neuron, fill=blue1] (X1) at (\xinput,2*\ydist) {$x^{(1)}$};
		\node[neuron, fill=blue1] (X2) at (\xinput,1*\ydist) {$x^{(2)}$};
		\node at (\xinput,0) {$\vdots$};
		\node[neuron, fill=blue1] (Xn) at (\xinput,-1*\ydist) {$x^{(n)}$};
		
		\node[neuron, fill=blue2] (H1) at (\xhidden,2*\ydist) {$h^{(1)}$};
		\node[neuron, fill=blue2] (H2) at (\xhidden,1*\ydist) {$h^{(2)}$};
		\node at (\xhidden,0) {$\vdots$};
		\node[neuron, fill=blue2] (Hn) at (\xhidden,-1*\ydist) {$h^{(n)}$};
		
		\node[neuron, fill=blue3] (Z1) at (\xoutput,2*\ydist) {$z^{(1)}$};
		\node[neuron, fill=blue3] (Z2) at (\xoutput,1*\ydist) {$z^{(2)}$};
		\node at (\xoutput,0) {$\vdots$};
		\node[neuron, fill=blue3] (Zn) at (\xoutput,-1*\ydist) {$z^{(n)}$};
		
		\node[lzcbox, fill=blue4!20] (parser) at (\xlzc,2.3) {$z^{(i)} \mapsto \{s_1, s_2, \dots, s_k\}$};
		\node[lzcbox, fill=blue4!30] (dict) at (\xlzc,0.7) {$\mathcal{D} = \{s_j\}_{j=1}^k$};
		\node[lzcbox, fill=blue5!30] (counter) at (\xlzc,-1.0) {$C(n) = |\mathcal{D}|$};
		
		\draw[arr] (counter.east) -- ++(0.5,0) node[right, font=\scriptsize] {\textbf{$\mathrm{LZC}(\mathbf{z}^{(i)}) \in \mathbb{R}$}};
		
		\foreach \i in {1,2,n} {
			\foreach \j in {1,2,n} {
				\draw[arr] (X\i) -- (H\j);
				\draw[arr] (H\i) -- (Z\j);
			}
		}
		
		\draw[arr] (Z1.east) -- (parser.west);
		\draw[arr] (Z2.east) -- ($(parser.west)+(0,0.1)$);
		\draw[arr] (Zn.east) -- ($(parser.west)+(0,-0.1)$);
		
		\draw[arr] (parser.south) -- (dict.north);
		\draw[arr] (dict.south) -- (counter.north);
	\end{tikzpicture}
	\label{fig:net}
	\caption{
		Schematic of the spiking neural network and its post-processing through Lempel–Ziv Complexity. The architecture includes an input layer $X = \{x^{(i)}\}$, hidden layer $H = \{h^{(i)}\}$, and output layer $Z = \{z^{(i)}\}$, each consisting of $n$ neurons. Binary input sequences of length 1024 are encoded into 
		$n$-bit spike trains and propagated through the network. The output activity of each neuron $z^{(i)}$ is converted into a binary sequence $\mathbf{x}_n = [x_1, x_2, \dots, x_n]$, which is parsed into substrings to build a dictionary $\mathcal{D}$ of unique patterns. The LZC is computed as $c_{\alpha}(\mathbf{x}_n) = 
		\frac{C_{\alpha}(\mathbf{x}_n)}{{n}} \log_{\alpha} n$, where $C_{\alpha}(\cdot)$ counts the number of distinct substrings that appear consecutively along the sequence. This measure quantifies the spatiotemporal complexity of the spike output.
	}
\end{figure*}
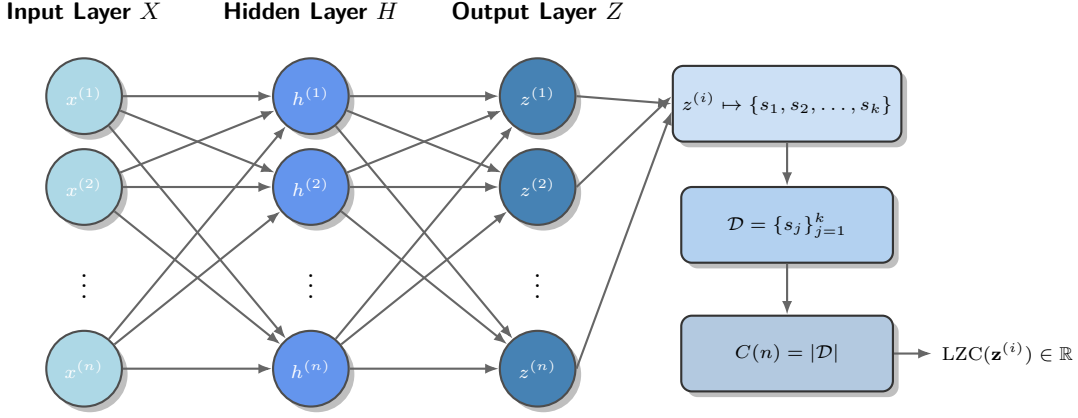
\par
The network architecture under consideration is made up of three distinct neuronal layers: input, hidden, and output.
Each containing $n$ computational units, as illustrated in Figure~\ref{fig:net}. The input layer receives binary sequences of length 1024, which are encoded in spike-train representations of $n$-bit width and processed through the network. The spike trains produced by the output neurons $\{z^{(i)}\}$ are decoded into binary sequences  $\mathbf{x}_n = [x_1, x_2, \dots, x_n]$, which are analyzed using the Lempel–Ziv Complexity metric. The LZC measures the number of unique substrings $C_{\alpha}(\mathbf{x}_n)$ encountered during sequential parsing. 
To normalize this measure across sequence lengths, we compute
$c_{\alpha}(\mathbf{x}_n) = \frac{C_{\alpha}(\mathbf{x}_n)}{{n}} \log_{\alpha} n$, where $\alpha = 2$ for binary sequences \cite{LempelZiv1976}. This provides a compact descriptor of the structural complexity of the network’s output activity. This hybrid approach combines the temporal precision and biological plausibility of SNNs with the capacity of LZC to quantify the structural complexity of spike patterns. By leveraging these complementary properties, the model enables efficient and interpretable classification of spatiotemporal neural data \cite{LempelZiv1976, Arnold2013, Pregowska2019}. This integration also enhances robustness to noise and improves performance, particularly when processing signals with variable temporal dynamics, such as those generated by Poisson processes \cite{Kara2000,Barbieri2001,Averbeck2009,Maric2025}.

\par
Poisson spike trains are produced using a homogeneous Poisson process, characterized by the independent occurrence of events at a fixed mean rate $\lambda$. The probability of observing a spike within a small time interval $\Delta t$ is given by $P (\lambda \Delta t) = o(\Delta t)$. The inter-spike intervals (ISIs) follow an exponential distribution $P (\lambda \Delta t) = \lambda e^{-\lambda \Delta t}$. Thus, spike sequences are generated with specific rates $\lambda_{x}$ and $\lambda_{y}$ for different input channels. To assess the influence of temporal variability on network dynamics and classification accuracy, these spike trains are evaluated under multiple rate conditions. The binary sequence of each process is converted into inputs to neuron models with specified $\Delta t = 1 ms$.

\section{Learning Rules}\label{learru}

Backpropagation (BP) is a gradient-based learning algorithm used to train neural networks by minimizing a loss function $ \mathcal{E} $ \cite{Singh2022}. The network parameters are updated through weight update

\[
\Delta w_i = -\eta \frac{\partial \mathcal{E}}{\partial w_i}, \quad
\frac{\partial \mathcal{E}}{\partial w_i} = 
\frac{\partial \mathcal{E}}{\partial V_m} \cdot \frac{\partial V_m}{\partial w_i},
\]
where $\eta$ is the learning rate and \( V_m \) denotes an intermediate potential. BP propagates error gradients backward through the network to optimize weights \cite{Kaur2022}, see Figure \ref{fig:snn_bp}.

\begin{figure*}[ht]
	\centering
	\begin{tikzpicture}[
		neuron/.style={circle, draw=black!70, thick, minimum size=1cm, inner sep=0pt, drop shadow, font=\scriptsize\bfseries, text=white},
		layerlabel/.style={font=\bfseries\small\sffamily},
		arr/.style={-{Latex[length=1.8mm]}, thick, draw=blue!80},
		backarr/.style={-{Latex[length=1.8mm]}, thick, dashed, draw=red!70!black},
		lzcbox/.style={rectangle, draw=black!80, thick, rounded corners=3pt, minimum width=2.8cm, minimum height=1cm, font=\scriptsize, align=center, fill=blue!5, drop shadow},
		classifier/.style={rectangle, draw=black, thick, rounded corners=3pt, minimum width=2.4cm, minimum height=0.8cm, font=\scriptsize, fill=green!10},
		legend/.style={font=\tiny\itshape}
		]
		
		\def\xinput{0}
		\def\xhidden{3}
		\def\xoutput{6}
		\def\xlzc{9.5}
		\def\ydist{1.2}
		
		\node[layerlabel] at (\xinput,3.5) {Input Layer $X$};
		\node[layerlabel] at (\xhidden,3.5) {Hidden Layer $H$};
		\node[layerlabel] at (\xoutput,3.5) {Output Layer $Z$};
		
		\node[neuron, fill=blue!20] (X1) at (\xinput,2*\ydist) {$x^{(1)}$};
		\node[neuron, fill=blue!20] (X2) at (\xinput,1*\ydist) {$x^{(2)}$};
		\node at (\xinput,0) {\vdots};
		\node[neuron, fill=blue!20] (Xn) at (\xinput,-1*\ydist) {$x^{(n)}$};
		
		\node[neuron, fill=blue!35] (H1) at (\xhidden,2*\ydist) {$h^{(1)}$};
		\node[neuron, fill=blue!35] (H2) at (\xhidden,1*\ydist) {$h^{(2)}$};
		\node at (\xhidden,0) {\vdots};
		\node[neuron, fill=blue!35] (Hn) at (\xhidden,-1*\ydist) {$h^{(n)}$};
		
		\node[neuron, fill=blue!50] (Z1) at (\xoutput,2*\ydist) {$z^{(1)}$};
		\node[neuron, fill=blue!50] (Z2) at (\xoutput,1*\ydist) {$z^{(2)}$};
		\node at (\xoutput,0) {\vdots};
		\node[neuron, fill=blue!50] (Zn) at (\xoutput,-1*\ydist) {$z^{(n)}$};
		
		\node[lzcbox] (parser) at (\xlzc,2.3) {$z^{(i)} \mapsto \{s_1, \dots, s_k\}$};
		\node[lzcbox] (dict) at (\xlzc,0.7) {$\mathcal{D} = \{s_j\}$};
		\node[lzcbox] (counter) at (\xlzc,-1.0) {$C(n) = |\mathcal{D}|$};
		\node[classifier] (classify) at (\xlzc+3.2,-1.0) {Class Decision};
		
		\foreach \i in {1,2,n} {
			\foreach \j in {1,2,n} {
				\draw[arr] (X\i) -- (H\j);
				\draw[arr] (H\i) -- (Z\j);
			}
		}
		\draw[arr] (Z1.east) -- (parser.west);
		\draw[arr] (Z2.east) -- ($(parser.west)+(0,0.1)$);
		\draw[arr] (Zn.east) -- ($(parser.west)+(0,-0.1)$);
		\draw[arr] (parser.south) -- (dict.north);
		\draw[arr] (dict.south) -- (counter.north);
		\draw[arr] (counter.east) -- (classify.west);
		
		\foreach \j in {1,2,n} {
			\foreach \k in {1,2,n} {
				\draw[backarr] (Z\j) -- (H\k);
				\draw[backarr] (H\j) -- (X\k);
			}
		}
		
		\node[legend] at (3.8,3.3) {\textcolor{blue!80}{Feedforward signal}};
		\draw[arr] (3.5,3.2) -- ++(1.2,0);
		
		\node[legend] at (3.8,2.9) {\textcolor{red!70!black}{Backpropagation (global)}};
		\draw[backarr] (3.5,2.8) -- ++(1.2,0);
		
	\end{tikzpicture}
	\caption{SNN trained with backpropagation learning. Output spike trains $\mathbf{z}^{(i)}$ are analyzed by LZC for classification. Error signals (red) propagate globally from output to input.}
	\label{fig:snn_bp}
\end{figure*}
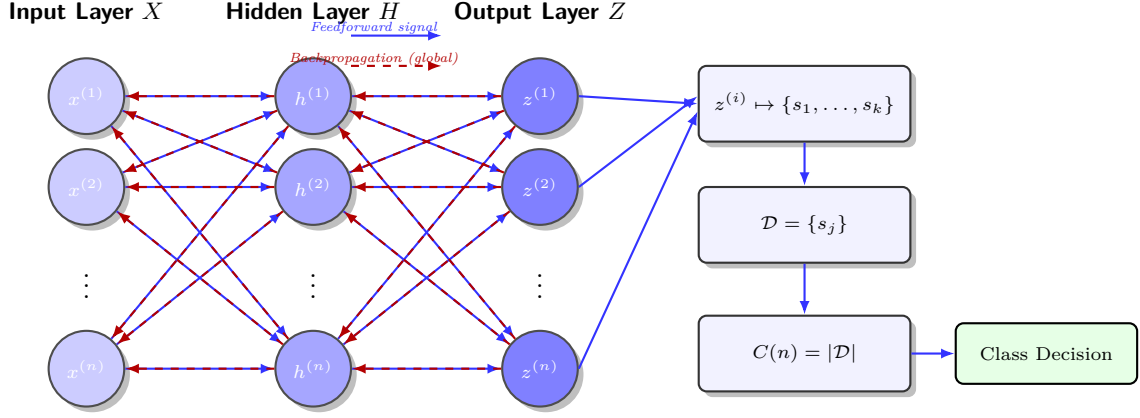

\begin{figure*}[ht]
	\centering
	\begin{tikzpicture}[
		neuron/.style={circle, draw=black!70, thick, minimum size=1cm, inner sep=0pt, drop shadow, font=\scriptsize\bfseries, text=white},
		layerlabel/.style={font=\bfseries\small\sffamily},
		arr/.style={-{Latex[length=1.8mm]}, thick, draw=blue!80},
		stdparr/.style={-{Latex[length=1.5mm]}, thick, dashed, draw=purple!80!black},
		lzcbox/.style={rectangle, draw=black!80, thick, rounded corners=3pt, minimum width=2.8cm, minimum height=1cm, font=\scriptsize, align=center, fill=blue!5, drop shadow},
		classifier/.style={rectangle, draw=black, thick, rounded corners=3pt, minimum width=2.4cm, minimum height=0.8cm, font=\scriptsize, fill=green!10},
		legend/.style={font=\tiny\itshape}
		]
		
		\def\xinput{0}
		\def\xhidden{3}
		\def\xoutput{6}
		\def\xlzc{9.5}
		\def\ydist{1.2}
		
		\node[layerlabel] at (\xinput,3.5) {Input Layer $X$};
		\node[layerlabel] at (\xhidden,3.5) {Hidden Layer $H$};
		\node[layerlabel] at (\xoutput,3.5) {Output Layer $Z$};
		
		\node[neuron, fill=blue!20] (X1) at (\xinput,2*\ydist) {$x^{(1)}$};
		\node[neuron, fill=blue!20] (X2) at (\xinput,1*\ydist) {$x^{(2)}$};
		\node at (\xinput,0) {\vdots};
		\node[neuron, fill=blue!20] (Xn) at (\xinput,-1*\ydist) {$x^{(n)}$};
		
		\node[neuron, fill=blue!35] (H1) at (\xhidden,2*\ydist) {$h^{(1)}$};
		\node[neuron, fill=blue!35] (H2) at (\xhidden,1*\ydist) {$h^{(2)}$};
		\node at (\xhidden,0) {\vdots};
		\node[neuron, fill=blue!35] (Hn) at (\xhidden,-1*\ydist) {$h^{(n)}$};
		
		\node[neuron, fill=blue!50] (Z1) at (\xoutput,2*\ydist) {$z^{(1)}$};
		\node[neuron, fill=blue!50] (Z2) at (\xoutput,1*\ydist) {$z^{(2)}$};
		\node at (\xoutput,0) {\vdots};
		\node[neuron, fill=blue!50] (Zn) at (\xoutput,-1*\ydist) {$z^{(n)}$};
		
		\node[lzcbox] (parser) at (\xlzc,2.3) {$z^{(i)} \mapsto \{s_1, \dots, s_k\}$};
		\node[lzcbox] (dict) at (\xlzc,0.7) {$\mathcal{D} = \{s_j\}$};
		\node[lzcbox] (counter) at (\xlzc,-1.0) {$C(n) = |\mathcal{D}|$};
		\node[classifier] (classify) at (\xlzc+3.2,-1.0) {Class Decision};
		
		\foreach \i in {1,2,n} {
			\foreach \j in {1,2,n} {
				\draw[arr] (X\i) -- (H\j);
				\draw[arr] (H\i) -- (Z\j);
			}
		}
		\draw[arr] (Z1.east) -- (parser.west);
		\draw[arr] (Z2.east) -- ($(parser.west)+(0,0.1)$);
		\draw[arr] (Zn.east) -- ($(parser.west)+(0,-0.1)$);
		\draw[arr] (parser.south) -- (dict.north);
		\draw[arr] (dict.south) -- (counter.north);
		\draw[arr] (counter.east) -- (classify.west);
		
		\foreach \i in {1,2,n} {
			\foreach \j in {1,2,n} {
				\draw[stdparr] (X\i) to[bend right=15] (H\j);
				\draw[stdparr] (H\i) to[bend right=15] (Z\j);
			}
		}
		
		\node[legend] at (3.8,3.3) {\textcolor{blue!80}{Feedforward signal}};
		\draw[arr] (3.5,3.2) -- ++(1.2,0);
		
		\node[legend] at (3.8,2.9) {\textcolor{purple!80!black}{STDP (local timing-based update)}};
		\draw[stdparr] (3.5,2.8) -- ++(1.2,0);

	\end{tikzpicture}
	\caption{SNN trained using STDP. Local spike timing between pre- and post-synaptic neurons adjusts weights. }
	\label{fig:snn_stdp_lzc}
\end{figure*}
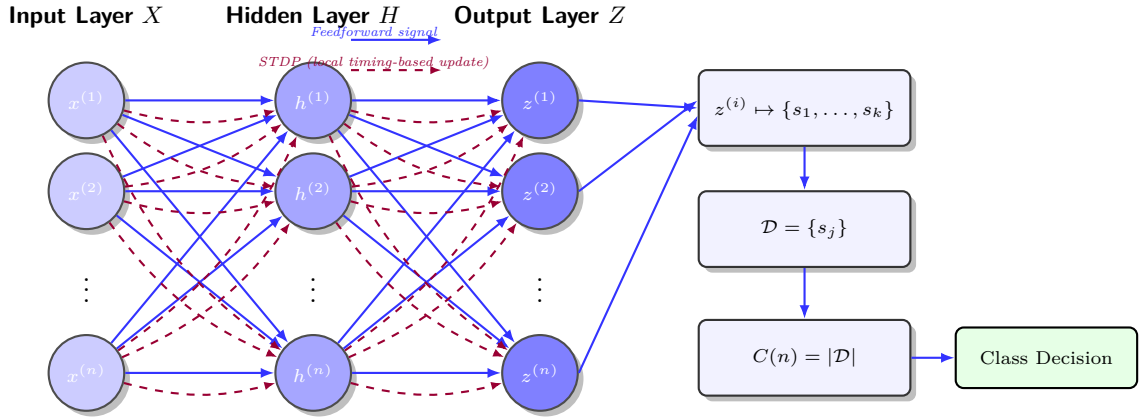

On the other hand, STDP is a biologically grounded, unsupervised learning mechanism in which synaptic weight changes depend on the precise relative timing of presynaptic and postsynaptic spikes \cite{Shah2019}, see Figure \ref{fig:snn_stdp_lzc}. When a presynaptic spike precedes a postsynaptic spike within a critical window, the synapse is potentiated (long-term potentiation, LTP); if the order is reversed, it undergoes depression (long-term depression, LTD). The weight update is defined as \cite{Markram2011,Merolla2014,Lagani2023}: 
\[
\Delta w_i =
\begin{cases}
	A_{+}\exp\!\Big(-\frac{t_{\mathrm{post}} - t_{\mathrm{pre}}}{\tau_{+}}\Big), & t_{\mathrm{post}} > t_{\mathrm{pre}}, \\[6pt]
	-A_{-}\exp\!\Big(-\frac{t_{\mathrm{pre}} - t_{\mathrm{post}}}{\tau_{-}}\Big), & t_{\mathrm{pre}} > t_{\mathrm{post}},
\end{cases}
\]
where $A_{\pm}$ are amplitudes of potentiation/depression and $\tau_{\pm}$ are their time constants.

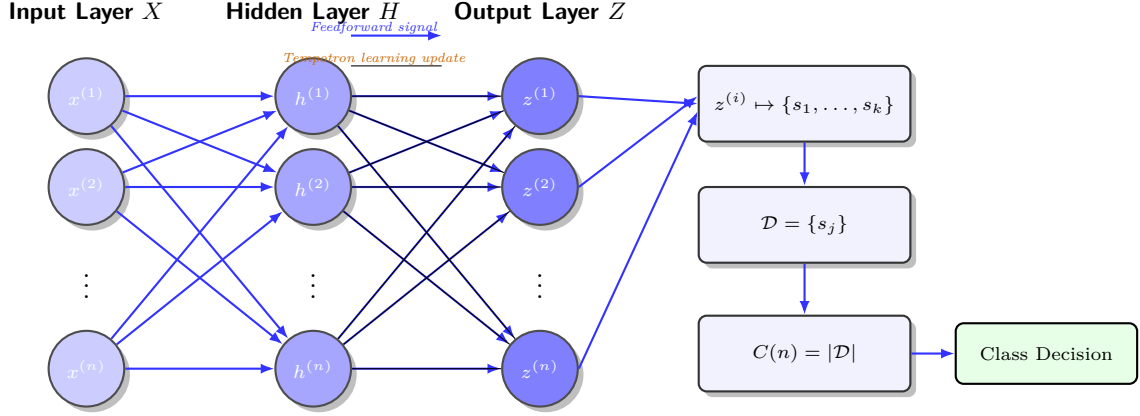
\begin{figure*}[h]
	\centering
	\begin{tikzpicture}[
		neuron/.style={circle, draw=black!70, thick, minimum size=1cm, inner sep=0pt, drop shadow, font=\scriptsize\bfseries, text=white},
		layerlabel/.style={font=\bfseries\small\sffamily},
		arr/.style={-{Latex[length=1.8mm]}, thick, draw=blue!80},
		tempopath/.style={-{Latex[length=1.8mm]}, thick, dashed, draw=orange!80!black}
		lzcbox/.style={rectangle, draw=black!80, thick, rounded corners=3pt, minimum width=2.8cm, minimum height=1cm, font=\scriptsize, align=center, fill=blue!5, drop shadow},
		classifier/.style={rectangle, draw=black, thick, rounded corners=3pt, minimum width=2.4cm, minimum height=0.8cm, font=\scriptsize, fill=green!10},
		legend/.style={font=\tiny\itshape}
		]
		
		\def\xinput{0}
		\def\xhidden{3}
		\def\xoutput{6}
		\def\xlzc{9.5}
		\def\ydist{1.2}
		
		\node[layerlabel] at (\xinput,3.5) {Input Layer $X$};
		\node[layerlabel] at (\xhidden,3.5) {Hidden Layer $H$};
		\node[layerlabel] at (\xoutput,3.5) {Output Layer $Z$};
		
		\node[neuron, fill=blue!20] (X1) at (\xinput,2*\ydist) {$x^{(1)}$};
		\node[neuron, fill=blue!20] (X2) at (\xinput,1*\ydist) {$x^{(2)}$};
		\node at (\xinput,0) {\vdots};
		\node[neuron, fill=blue!20] (Xn) at (\xinput,-1*\ydist) {$x^{(n)}$};
		
		\node[neuron, fill=blue!35] (H1) at (\xhidden,2*\ydist) {$h^{(1)}$};
		\node[neuron, fill=blue!35] (H2) at (\xhidden,1*\ydist) {$h^{(2)}$};
		\node at (\xhidden,0) {\vdots};
		\node[neuron, fill=blue!35] (Hn) at (\xhidden,-1*\ydist) {$h^{(n)}$};
		
		\node[neuron, fill=blue!50] (Z1) at (\xoutput,2*\ydist) {$z^{(1)}$};
		\node[neuron, fill=blue!50] (Z2) at (\xoutput,1*\ydist) {$z^{(2)}$};
		\node at (\xoutput,0) {\vdots};
		\node[neuron, fill=blue!50] (Zn) at (\xoutput,-1*\ydist) {$z^{(n)}$};
		
		\node[lzcbox] (parser) at (\xlzc,2.3) {$z^{(i)} \mapsto \{s_1, \dots, s_k\}$};
		\node[lzcbox] (dict) at (\xlzc,0.7) {$\mathcal{D} = \{s_j\}$};
		\node[lzcbox] (counter) at (\xlzc,-1.0) {$C(n) = |\mathcal{D}|$};
		\node[classifier] (classify) at (\xlzc+3.2,-1.0) {Class Decision};
		
		\foreach \i in {1,2,n} {
			\foreach \j in {1,2,n} {
				\draw[arr] (X\i) -- (H\j);
				\draw[arr] (H\i) -- (Z\j);
			}
		}
		\draw[arr] (Z1.east) -- (parser.west);
		\draw[arr] (Z2.east) -- ($(parser.west)+(0,0.1)$);
		\draw[arr] (Zn.east) -- ($(parser.west)+(0,-0.1)$);
		\draw[arr] (parser.south) -- (dict.north);
		\draw[arr] (dict.south) -- (counter.north);
		\draw[arr] (counter.east) -- (classify.west);
		
		\foreach \j in {1,2,n} {
			\foreach \k in {1,2,n} {
				\draw[tempopath] (Z\j) -- (H\k);
			}
		}
		
		\node[legend] at (3.8,3.3) {\textcolor{blue!80}{Feedforward signal}};
		\draw[arr] (3.5,3.2) -- ++(1.2,0);
		
		\node[legend] at (3.8,2.9) {\textcolor{orange!80!black}{Tempotron learning update}};
		\draw[tempopath] (3.5,2.8) -- ++(1.2,0);

	\end{tikzpicture}
	\caption{SNN  trained using Tempotron rule. Local error signals flow from the output to hidden layer.}
	\label{fig:snn_tempotron_lzc}
\end{figure*}

The Tempotron is a biologically inspired supervised learning rule designed for SNNs, which emphasizes temporal precision over spike rate \cite{Gutig2006}. At its core, it uses LIF neurons to integrate time-varying postsynaptic potentials (PSPs) from presynaptic spikes, see Figure \ref{fig:snn_tempotron_lzc}. If the neuron should fire ($P^{+}$) or stay silent ($P^{-}$), weights are updated as:
\[
\Delta w_i = \eta \sum_{t \in T_{\mathrm{pre}}} K(t_f - t),
\] 
where $T_{\mathrm{pre}}$ is the set of presynaptic spike times, $K(\cdot)$ is the postsynaptic potential kernel, $t_f$ is the time of maximal membrane potential, and $\eta$ is the learning rate.
This corresponds to gradient descent on the peak membrane potential $V_m(t)$. This learning rule performs gradient descent on the peak membrane potential to minimize classification error. Compared to traditional rate-based approaches or STDP, the Tempotron requires fewer tunable parameters, operates under supervised feedback, and is optimally designed for temporal pattern recognition tasks where the timing of spikes conveys critical information.

\section{Results} 
In our study, we introduce a novel spiking neural network framework that, to the best of our knowledge, is the first to integrate Lempel–Ziv Complexity with biologically inspired neuron models. We employ two distinct SNN architectures: one based on the widely used leaky integrate-and-fire neurons \cite{Dutta2017}, and the other utilizing a biologically motivated meta neuron model \cite{Yao2024}. This integration of LZC with SNNs provides a new classification approach, leveraging the temporal coding capabilities and neural plausibility of SNNs along with the quantification of structural complexity provided by LZC. The result is a robust and interpretable method for analyzing spatio-temporal data, an aspect that remains largely unexplored in the current SNN literature. To optimize performance and evaluate learning dynamics, we implement and compare several training algorithms, including backpropagation-based methods \cite{Wu2018}, spike-timing-dependent plasticity \cite{Hao2020}, and the Tempotron learning rule \cite{Gutig2006}.

\par
Table~\ref{tab:accuracy_1st_db} presents the classification accuracy results for spiking neural networks based on the leaky integrate-and-fire neuron model, using inputs generated as Poisson spike trains. Our findings indicate that optimizing both learning parameters and intrinsic neuron parameters can significantly improve classification accuracy. In particular, the application of backpropagation and spike-timing-dependent plasticity leads to gains from 82.50\% to 96.00\% (13.50 percentage points) and from 86.50\% to 96.50\% (10.00 percentage points), respectively, when thresholds and decay constants are learned in addition to the number of neurons and the learning rate. In contrast, the Tempotron learning rule exhibits a more modest improvement from 96.00\% to 99.50\% (3.50 percentage points). However, its baseline performance, achieved by training only the number of neurons and the learning rate, is already high, reaching 96.00\% accuracy.



\begin{table*}[t]
	\centering
	\scriptsize
	\caption{Accuracy results for SNN based on LIF neuron model.}
	\label{tab:accuracy_1st_db}
	\resizebox{\textwidth}{!}{%
		\begin{tabular}{lllccccccc}
			\toprule
			\textbf{Learning algorithm} & \textbf{Input} & \textbf{Learning parameters} 
			& \textbf{Acc [\%]} & \textbf{P$_0$} & \textbf{R$_0$} & \textbf{F1$_0$} 
			& \textbf{P$_1$} & \textbf{R$_1$} & \textbf{F1$_1$} \\
			\midrule
			BP        & Poisson & \makecell[l]{Number of neurons\\Learning rate\\Thresholds\\Decays} 
			& 96.00($13.50 \uparrow$)& 0.6809 & 0.9796 & 0.8033 & 0.9661  & 0.5588 & 0.7081 \\
			\\
			&  & \makecell[l]{Number of neurons\\Learning rate} 
			& 82.50 & 1.0000 & 0.0510 & 0.0971 & 0.5231 & 1.0000 & 0.6869 \\
			
			\midrule
			STDP      & Poisson & \makecell[l]{Number of neurons\\Learning rate\\Thresholds\\Decays} 
			& 96.50($10.00\uparrow$)& 0.9333 & 1.0000 & 0.9655 & 1.0000 & 0.9314 & 0.9645 \\
			\\
			&  & \makecell[l]{Number of neurons\\Learning rate} 
			&  86.50& 0.8515 & 0.8776 & 0.8643 & 0.8799 & 0.8529 & 0.8657 \\
			
			\midrule
			Tempotron & Poisson & \makecell[l]{Number of neurons\\Learning rate\\Thresholds\\Decays} 
			& 99.50 (3.50$\uparrow$) & 1.0000 & 0.0204 & 0.0400 & 0.5152 & 1.0000 & 0.6800 \\
			\\
			&  & \makecell[l]{Number of neurons\\Learning rate} 
			&  96.00& 1.0000 & 0.0102 & 0.0202 & 0.5126 & 1.0000 & 0.6777 \\
			\bottomrule
		\end{tabular}%
	}
	\smallskip
	\textit{Notes.} 
	P\textsubscript{0}, R\textsubscript{0}, F1\textsubscript{0} denote Precision, Recall, and F1-score for Class 0, respectively. 
	Analogously, P\textsubscript{1}, R\textsubscript{1}, F1\textsubscript{1} correspond to Precision, Recall, and F1-score for Class 1.
\end{table*}

\begin{table*}[t]
	\centering
	\scriptsize
	\caption{Comparison of computational cost and time efficiency of the evaluated SNN based on LIF neuron model.}
	\label{tab:efficiency_metrics_1}
	\resizebox{\textwidth}{!}{%
		\begin{tabular}{lllccccccccc}
			\toprule
			\textbf{Learning algorithm} & \textbf{Input} & \textbf{Learning Rules}& $\overline{t}$ [ms] & $t_{95}$ [ms] & \textbf{Throughput [samples/s]} & \textbf{FLOPs fwd} & \textbf{FLOPs bwd} & 
			\textbf{Total FLOPs} & \textbf{Energy [J]} & \textbf{Train FLOPs/s} & \textbf{Eval FLOPs/s}\\
			\midrule
			BP  & Poisson & \makecell[l]{Number of neurons\\Learning rate\\Thresholds\\Decays} & 0.3672 & 0.9249 & 2677.16 & 25600 & 25600 
			& 926\,720\,000 & 0.000927 & 1.36e+06 & 6.85e+07 \\ \\
			\\
			&  & \makecell[l]{Number of neurons\\Learning rate}       & 0.5663 & 1.2282 & 1731.02 & 50176 & 50176
			& 1\,816\,371\,200 & 0.001816 & 1.35e+06 & 8.69e+07  \\
			\midrule
			STDP  & Poisson & \makecell[l]{Number of neurons\\Learning rate\\Thresholds\\Decays}   & 0.7678 & 1.2600 & 1279.03 & 65536 & 65536
			& 1\,192\,755\,200 & 0.001193 & 1.42e+07 & 8.38e+07  \\
			\\
			&  & \makecell[l]{Number of neurons\\Learning rate}         & 0.8998 & 1.1933 & 1102.03 & 1568 & 1568
			& 56\,761\,600 & 0.000057 & 4.94e+05 & 1.73e+06   \\
			\midrule
			Tempotron & Poisson & \makecell[l]{Number of neurons\\Learning rate\\Thresholds\\Decays}    & 0.1584 & 0.2333 & 6080.64 & 4096 & 4096
			& 148\,275\,200 & 0.000148 & 3.32e+07 & 2.49e+07 \\
			\\
			&  & \makecell[l]{Number of neurons\\Learning rate}         & 0.1970 & 0.5060 & 4702.25 & 4096 & 4096
			& 148\,275\,200 & 0.000148 & 3.48e+07 & 1.93e+07  \\
			\bottomrule
		\end{tabular}%
	}
\end{table*}

Table~\ref{tab:efficiency_metrics_1} summarizes the computational cost and time efficiency of the LIF-based SNN for different learning rules and configurations. Across all methods, inference remains in the sub-millisecond regime, with average latency per sample ranging from $0.16$\,ms to $0.90$\,ms and 95th percentile latency below $1.3$\,ms. This corresponds to a throughput between approximately $1.1\times 10^{3}$ and $6.1\times 10^{3}$ samples/s, indicating that all configurations are suitable for real-time processing. The extended configurations, in which intrinsic neuron parameters are learned in addition to synaptic weights, do not systematically increase the computational burden. For backpropagation, the extended model achieves a lower average latency ($0.37$\,ms \textit{versus}\ $0.57$\,ms) and higher throughput (2677\,samples/s vs.\ 1731\,samples/s) despite using fewer FLOPs overall ( $9.27\times 10^{8}$ \textit{versus} $1.82\times 10^{9}$). In contrast, the extended STDP configuration requires more compute per sample (up to 65\,536 FLOPs in forward and backward passes) and a higher total FLOP budget than its baseline, but still maintains sub-millisecond latency. Tempotron-based LIF-SNNs achieve the highest throughput, with the extended variant reaching on average $0.16$\,ms per sample and over 6000\,samples/s, at a moderate total cost of $1.48\times 10^{8}$ FLOPs. Overall, the results show that substantial accuracy gains (Table~\ref{tab:accuracy_1st_db}) can be obtained without sacrificing real-time performance, and in some cases with improved speed--accuracy trade-offs.


In Table~\ref{tab:accuracy_2st_db}, the classification accuracy results for spiking neural networks based on the meta-neuron model are reported. Similar to the LIF-based architecture, jointly optimizing both intrinsic neuron parameters and the learning rate leads to a clear improvement in classification accuracy. Among the evaluated learning algorithms, backpropagation yields the largest gain, increasing accuracy from 89.00\% in the baseline configuration (number of neurons and learning rate only) to 97.50\% when thresholds and decay constants are also learned, i.e., by 8.50 percentage points. STDP shows a comparable trend, with an improvement from 88.50\% to 95.50\% (7.00 percentage points), while Tempotron achieves a more moderate gain from 89.50\% to 93.00\% (3.50 percentage ponits).

\par

\begin{table*}[t]
	\centering
	\scriptsize
	\caption{Accuracy results for SNN based on meta neuron model.}
	\label{tab:accuracy_2st_db}
	\resizebox{\textwidth}{!}{%
		\begin{tabular}{lllccccccc}
			\toprule
			\textbf{Learning algorithm} & \textbf{Input} & \textbf{Learning parameters} 
			& \textbf{Acc [\%]} & \textbf{P$_0$} & \textbf{R$_0$} & \textbf{F1$_0$} 
			& \textbf{P$_1$} & \textbf{R$_1$} & \textbf{F1$_1$} \\
			\midrule
			BP        & Poisson & \makecell[l]{Number of neurons\\Learning rate\\Thresholds\\Decays} 
			& 97.50 (8.50$\uparrow$)& 0.9896 & 0.9694 & 0.9794 & 0.9712 & 0.9902 & 0.9806 \\
			\\
			&  & \makecell[l]{Number of neurons\\Learning rate} 
			& 89.00 & 0.4900 & 1.0000 & 0.6577 & 0.0000 & 0.0000 & 0.0000 \\
			
			\midrule
			STDP      & Poisson & \makecell[l]{Number of neurons\\Learning rate\\Thresholds\\Decays} 
			& 95.50 (8.00$\uparrow$)& 0.4900 & 1.0000 & 0.6577 & 0.0000 & 0.0000 & 0.0000 \\
			\\
			&  & \makecell[l]{Number of neurons\\Learning rate} 
			& 88.50 & 0.4900 & 1.0000 & 0.6577 & 0.0000 & 0.0000 & 0.0000 \\
			
			\midrule
			Tempotron & Poisson & \makecell[l]{Number of neurons\\Learning rate\\Thresholds\\Decays} 
			&93.00 (3.50 $\uparrow$) & 0.9468 & 0.9082 & 0.9271 & 0.9151 & 0.9510 & 0.9327 \\
			\\
			&  & \makecell[l]{Number of neurons\\Learning rate} 
			& 89.50 & 0.7778 & 0.9286 & 0.8465 & 0.9157 & 0.7451 & 0.8216 \\
			\bottomrule
		\end{tabular}%
	}
	\smallskip
	\textit{Notes.} 
	P\textsubscript{0}, R\textsubscript{0}, F1\textsubscript{0} denote Precision, Recall, and F1-score for Class 0, respectively. 
	Analogously, P\textsubscript{1}, R\textsubscript{1}, F1\textsubscript{1} correspond to Precision, Recall, and F1-score for Class 1.
\end{table*}

\begin{table*}[t]
	\centering
	\scriptsize
	\caption{Comparison of computational cost and time efficiency of the evaluated SNN based on meta neuron model.}
	\label{tab:efficiency_metrics}
	\resizebox{\textwidth}{!}{%
		\begin{tabular}{lllccccccccc}
			\toprule
			\textbf{Learning algorithm} & \textbf{Input} & \textbf{Learning Rules}& $\overline{t}$ [ms] & $t_{95}$ [ms] & \textbf{Throughput [samples/s]} & \textbf{FLOPs fwd} & \textbf{FLOPs bwd} & 
			\textbf{Total FLOPs} & \textbf{Energy [J]} & \textbf{Train FLOPs/s} & \textbf{Eval FLOPs/s}\\
			\midrule
			BP  & Poisson & \makecell[l]{Number of neurons\\Learning rate\\Thresholds\\Decays}  & 0.9466 & 1.6320 & 1046.74 & 1800 & 1800 & 65\,160\,000  & 0.000065 & 5.80e+05  & 1.88e+06 \\
			\\
			&  & \makecell[l]{Number of neurons\\Learning rate}              & 0.7720 & 1.6226 & 1265.20 & 1568 & 1568 & 56\,761\,600  & 0.000057 & 6.51e+05  & 1.98e+06 \\
			\midrule
			STDP  & Poisson & \makecell[l]{Number of neurons\\Learning rate\\Thresholds\\Decays}  & 0.7720 & 1.6226 & 1265.20 & 1568 & 1568 & 56\,761\,600  & 0.000057 & 6.51e+05  & 1.98e+06 \\
			\\
			&  & \makecell[l]{Number of neurons\\Learning rate}              & 0.7720 & 1.6226 & 1265.20 & 1568 & 1568 & 56\,761\,600  & 0.000057 & 6.51e+05  & 1.98e+06 \\
			\midrule
			Tempotron & Poisson & \makecell[l]{Number of neurons\\Learning rate\\Thresholds\\Decays}  & 0.6687 & 1.3717 & 1479.66 & 1568 & 1568 & 56\,761\,600  & 0.000057 & 4.28e+06 & 2.32e+06 \\
			\\
			&  & \makecell[l]{Number of neurons\\Learning rate}              & 0.5925 & 1.2652 & 1644.91 & 1568 & 1568 & 56 761 600 & 0.000057 & 4.43e+06 & 2.58e+06 \\
			\bottomrule
		\end{tabular}%
	}
\end{table*}

Table~\ref{tab:efficiency_metrics} reports the corresponding computational profile for the meta-neuron based SNN. In this case, the average inference latency lies between $0.59$\,ms and $0.95$\,ms, with 95th percentile latency remaining below $1.7$\,ms for all learning rules. The resulting throughput spans roughly from $1.0\times 10^{3}$ to $1.65\times 10^{3}$ samples/s, i.e., somewhat lower than the fastest LIF-based Tempotron configuration but still compatible with real-time scenarios. Unlike the LIF-SNN, the meta-SNN exhibits a more uniform computational footprint across learning rules: forward and backward passes typically require 1568--1800 FLOPs per sample, and the total FLOP budget stays on the order of $5.7$--$6.5\times 10^{7}$. Energy estimates remain in the $10^{-5}$\,J range under the assumed energy model. Within this regime, Tempotron again provides the most favourable time characteristics, achieving the lowest average latency (down to $0.59$\,ms) and the highest throughput (up to 1645\,samples/s), while backpropagation and STDP yield slightly higher latency at comparable compute cost. Taken together with the accuracy results in Table~\ref{tab:accuracy_2st_db}, these observations indicate that meta-neurons offer improved representational flexibility with only a modest overhead in runtime compared to standard LIF neurons, especially when combined with efficient learning rules such as Tempotron.

Across all configurations, inference remains in the sub-millisecond regime, with average latency per sample ranging from 0.16\,ms to 0.95\,ms and throughput between approximately $1.1\times 10^{3}$ and $6.1\times 10^{3}$ samples/s. While the extended configurations with learned intrinsic parameters increase the total number of FLOPs in some cases, they offer markedly higher accuracy, leading to a favourable trade-off between computational cost and classification performance. Notably, Tempotron-based models achieve the highest throughput, whereas BP- and STDP-based LIF-SNNs provide the best accuracy-to-compute ratio.

In addition to overall accuracy, Tables~\ref{tab:accuracy_1st_db} and \ref{tab:accuracy_2st_db} include class-wise precision, recall, and F1-scores, which provide a more detailed picture of model behaviour. The extended configurations, in which intrinsic neuron parameters are optimized jointly with synaptic weights, consistently lead to more balanced class-wise performance. This is reflected in higher and more symmetric F1-scores across classes, indicating that the models are not only more accurate overall but also less biased toward a particular class. In contrast, baseline configurations optimized only over network size and learning rate often exhibit strong class imbalance, with one class being detected almost perfectly (recall close to $\approx$ 1.0) while the other collapses to near-zero precision or recall. This behaviour is particularly evident for STDP and Tempotron baselines, where the model tends to favour the more frequent or more separable class. Learning intrinsic parameters alleviates this issue by enabling the neurons to adjust their internal timescales and thresholds, thereby improving separability in both classes. Overall, the class-wise metrics confirm that the accuracy improvements observed in the extended configurations stem not only from higher correct classification rates but from a more robust and balanced decision boundary across classes.

Two optimization configurations were compared for each learning algorithm: (1) one can say baseline, where the number of neurons and the learning rate were optimized, and (2) extended, where intrinsic neuron parameters (thresholds and decay constants) were additionally optimized. For both the LIF and meta-neuron architectures, adding trainable thresholds and decay constants consistently improved classification accuracy. This suggests that adjusting firing thresholds enhanced the neuron’s sensitivity to relevant spike patterns, while tuning decay constants adapted the temporal integration window of post-synaptic potentials. The combined effect allowed the network to more effectively capture temporal dependencies in the input. 

Backpropagation and STDP benefited the most from intrinsic parameter optimization, yielding accuracy gains of up to 13.50 percentage points and 10.00 percentage points over their respective baselines in the LIF-SNN, and up to 8.50 and 7.00 percentage points in the meta-SNN. Tempotron, while already showing a strong baseline performance (96.00\% for LIF-SNN and 89.50\% for meta-SNN), exhibited smaller improvements of about 3.50 percentage points. This suggests that its performance relies primarily on synaptic weight dynamics rather than intrinsic parameter tuning. As a result, parameter optimization (thresholds and decay constants) contributes less to performance for Tempotron compared to Backpropagation or STDP. Thus, the optimization configurations and their interpretation are summarized in Table~\ref{tab:config_interpretation}.

\begin{table*}[t]
	\centering
	\caption{Optimization configurations and their interpretation.}
	\label{tab:config_interpretation}
	\begin{tabular}{l p{6cm} p{6cm}}
		\toprule
		\textbf{Configuration} & \textbf{Optimized Parameters} & \textbf{Interpretation and Effect} \\
		\midrule
		Baseline & Number of neurons, learning rate & Focuses on global capacity and convergence speed. Provides stable but limited adaptation to temporal patterns. \\
		Extended & Number of neurons, learning rate, thresholds, decay constants & Adjusts firing sensitivity and temporal integration window. Increases responsiveness to relevant spike timings and improves accuracy, especially for Backpropagation and STDP. \\
		\bottomrule
	\end{tabular}
\end{table*}

\section{Numerical experiments details}

All experiments are conducted on sequences from two Poisson processes with random seeds applied generated by the system generator. Each sequence has length 1024, generating a total of 2000 sequences (1000 per class), after random shuffling and splitting them 90/10 into training and validation. We consider the intensity parameters: $\lambda_1 \in [0.5, 3.0],\quad \lambda_2 \in [1.5, 6.0], \text{with step size } 0.5, \text{with the constraint} \quad\lambda_2>\lambda_1, \Delta\lambda \in [0.5, 3.0]$.

\begin{table*}[t]
	\centering
	\caption{Search spaces for model configuration and neuron parameters, with ranges and discretization steps.}
	\begin{tabular}{lccccc}
		\toprule
		& \multicolumn{2}{c}{\textbf{LIF-SNN (50 trials)}} & \multicolumn{2}{c}{\textbf{Meta-SNN (200 trials)}} \\
		\cmidrule(lr){2-3}\cmidrule(lr){4-5}
		\textbf{Parameter} & \textbf{Range} & \textbf{Step} & \textbf{Range} & \textbf{Step} \\
		\midrule
		Number of neurons $N$ & $\{16,32,\dots,128\}$ & $16$ & $\{4,6,\dots,30\}$ & $2$ \\
		Number of sub-neurons $m$ & — & — & $\{2,4,6\}$ & $2$ \\
		Threshold $\theta^{(\mathrm{thr})}$ & $[0.1,\,1.0]$ & $0.1$ & $[0.5,\,2.0]$ & $0.1$ \\
		Leakage $\theta^{(\mathrm{decay})}$ & $[0.01,\,0.1]$ & $0.01$ & $[0.01,\,0.2]$ & $0.01$ \\
		Learning rate $lr$ & $[5{\times}10^{-4},\,10^{-1}]$ & log-uniform & $[10^{-3},\,10^{-1}]$ & $0.001$ \\
		\bottomrule
	\end{tabular}
	\label{hyp}
\end{table*}

We compare two fully connected SNN variants (LIF-SNN, Meta-SNN), each trained for 10 epochs. It is worth to stress that Meta-SNN extends the standard LIF-SNN by replacing each single neuron with a meta-neuron composed of $m$ parallel LIF sub-neurons. All sub-neurons share the same input, and their spike outputs are averaged into a single signal that represents the activity of the meta-neuron. This design introduces an additional degree of freedom, enabling richer spiking dynamics compared to conventional LIF-SNNs.

For LIF-SNN, each of the three layers has $n$ LIF neurons. Every neuron is defined by a decay parameter $\theta^{(\mathrm{decay})}$ and a firing threshold $\theta^{(\mathrm{thr})}$. Synaptic weights between layers are initialized with Xavier/Glorot normal distribution, $\mathcal{N}(0,\,1/n)$. Both weights and neuron parameters are trained jointly. After each update, neuron thresholds are clipped to $[0.1,\,1.0]$ and decays to $[0.01,\,0.10]$ to keep them in a biologically meaningful range. In our experiments, the layer width varies over $n\in\{16,32,\dots,128\}$, thresholds over $[0.1,1.0]$ (step $0.1$), decays over $[0.01,0.10]$ (step $0.01$), and the learning rate is drawn log-uniformly from $[5{\times}10^{-4},10^{-1}]$.

In Meta-SNN, each layer also contains $n$ units, but now each unit is a meta-neuron made up of $m$ parallel LIF sub-neurons. These sub-neurons receive the same input, and their spikes are aggregated to form the output of the meta-neuron. As in LIF-SNN, weights are Xavier-initialized and both synaptic and neuron parameters are trained jointly. After each update, meta-neuron thresholds are clipped to $[0.5,\,2.0]$ and decays to $[0.01,\,0.20]$. The ranges explored are $n\in\{4,6,\dots,30\}$ (step $2$) and $m\in\{2,4,6\}$, with thresholds and decays as above. The learning rate is scanned uniformly in $[10^{-3},10^{-1}]$ with step $0.001$. These bounds match the post-update clipping constraints and were kept fixed for all $(\lambda_1,\lambda_2)$ settings.

The parameters ranges, discretization steps, and default values for all learning rules are summarized in Table~\ref{tab:lranges}. Updates follow the mathematical definitions given in Sec.~\ref{learru}, with the PSP kernel normalized to $\max K = 1$. After each update, neuron parameters are clipped to 
$\theta^{(\mathrm{thr})} \in [0.1,\,2.0]$ 
and 
$\theta^{(\mathrm{decay})} \in [0.01,\,0.2]$.

\begin{table*}[t]
	\centering
	\caption{Parameter ranges for learning rules with discretization steps and defaults. All times in ms; $\Delta t=1$\,ms.}
	\resizebox{\textwidth}{!}{%
		\begin{tabular}{l l c c c}
			\toprule
			\textbf{Rule} & \textbf{Parameter} & \textbf{Range} & \textbf{Step / Scale} & \textbf{Default} \\
			\midrule
			\multirow{6}{*}{Tempotron}
			& $\eta$ (learning rate) & $[10^{-5},\,10^{-1}]$ & log-uniform & $10^{-4}$ \\
			& $\tau_m$ (membrane time constant) & $[10,\,30]$ & $5$ & $20$ \\
			& $\tau_s$ (synaptic time constant) & $[2,\,10]$ & $2$ & $5$ \\
			& $V_{\mathrm{th}}$ (spike threshold) & $[0.5,\,2.0]$ & $0.1$ & $0.5$ \\
			& $\|\Delta \mathbf{w}\|_{\infty}$ (weight bound) & $[10^{-3},\,1]$ & log-scale & $1$ \\
			& $T$ (sequence length) & $[512,\,1024]$ & $512$ & $1024$ \\
			\midrule
			\multirow{7}{*}{STDP}
			& $A_{+}$ (potentiation amplitude) & $[10^{-4},\,5\!\times\!10^{-2}]$ & log-uniform & $10^{-2}$ \\
			& $A_{-}$ (depression amplitude) & $[10^{-4},\,5\!\times\!10^{-2}]$ & log-uniform & $1.2\!\times\!10^{-2}$ \\
			& $\tau_{+}$ (time constant for LTP) & $[10,\,40]$ & $5$ & $20$ \\
			& $\tau_{-}$ (time constant for LTD) & $[20,\,60]$ & $5$ & $30$ \\
			& $s_{\mathrm{STDP}}$ (scaling factor) & $[0.1,\,1.0]$ & $0.1$ & $1.0$ \\
			& $W$ (time window size) & $[50,\,200]$ & $25$ & $100$ \\
			& $w_{\max}$ (max synaptic weight) & $[0.5,\,2.0]$ & $0.5$ & $1.0$ \\
			\midrule
			\multirow{6}{*}{BP}
			& $lr_0$ (initial learning rate) & $[10^{-4},\,10^{-1}]$ & log-uniform & $10^{-3}$ \\
			& $\gamma$ (LR decay factor) & $[0.95,\,1.00]$ & $0.01$ & $0.99$ \\
			& $k_s$ in $g_s(V){=}k_s e^{-V^2}$ (gain) & $[0.1,\,0.5]$ & $0.1$ & $0.3$ \\
			& $B$ (batch size) & $\{1\}$ & fixed & $1$ \\
			& $T$ (epochs) & $\{10\}$ & fixed & $10$ \\
			& $\mathcal{L}$ (loss function) & \{MSE, CE\} & choice & MSE (LIF), CE (Meta) \\
			\bottomrule
		\end{tabular}
	}
	\label{tab:lranges}
\end{table*}

All computations were run on an Intel(R) Core(TM) i7\textendash14700F (2.10\,GHz), Windows~10 (10.0.26100, x86\_64). Python~3.10.15 (conda\textendash forge); packages: NumPy~2.2.6, SciPy~1.14.1, Matplotlib~3.9.2, Seaborn~0.13.2, scikit\textendash learn~1.6.0, Optuna~4.1.0. NumPy was linked against SciPy\textendash OpenBLAS~0.3.29.

\section{Theoretical justification}

The empirical results in Tables~\ref{tab:accuracy_1st_db}--\ref{tab:efficiency_metrics}
ddemonstrate that jointly learning synaptic weights and intrinsic neuron parameters (such as firing thresholds and decay constants) leads to a substantial increase in classification accuracy for both LIF-based and meta-neuron-based SNNs. In this section, we provide a mathematical justification for this observation by explicitly linking neuron dynamics, learning algorithms, and achievable risk.

\paragraph{Problem setting}
Let $\mathcal{X}$ denote the space of input spike trains and let
$\mathcal{Y}=\{0,1\}$ be the label space.
The classifier implemented by the SNN combined with Lempel--Ziv complexity is defined as

\begin{equation}
	h_{w,\theta}(x)
	=
	g\!\Big(c_2\big(\Phi_{w,\theta}(x)\big)\Big),
	\qquad x\in\mathcal{X},
\end{equation}

where $\Phi_{w,\theta}$ denotes the spike-based transformation induced by the network,
parameterized by synaptic weights $w$ and intrinsic neuron parameters $\theta$,
$c_2(\cdot)$ is the normalized Lempel--Ziv complexity, and $g$ is a scalar decision function.
For a fixed data-generating distribution $(X,Y)\sim\mathcal{D}$ and loss function
$\ell:\mathcal{Y}\times\mathcal{Y}\to\mathbb{R}_+$, the expected risk is
\begin{equation}
	\widetilde{\mathcal{R}}(w,\theta)
	=
	\mathbb{E}_{(X,Y)\sim\mathcal{D}}
	\big[\ell(h_{w,\theta}(X),Y)\big].
\end{equation}

\paragraph{Baseline and extended hypothesis classes}
In the baseline configuration, intrinsic neuron parameters are fixed at
$\theta_0\in\Theta$, and only the synaptic weights are optimized. The corresponding
hypothesis class is
\begin{equation}
	\mathcal{H}_{\mathrm{base}}
	=
	\{h_{w,\theta_0}: w\in\mathcal{W}\}.
\end{equation}
In the extended configuration, both synaptic and intrinsic parameters are learnable,
leading to the larger class
\begin{equation}
	\mathcal{H}_{\mathrm{ext}}
	=
	\{h_{w,\theta}: (w,\theta)\in\mathcal{W}\times\Theta\}.
\end{equation}

\paragraph{Risk monotonicity under class enlargement}
Since $\theta_0\in\Theta$, it holds that
$\mathcal{H}_{\mathrm{base}}\subseteq\mathcal{H}_{\mathrm{ext}}$.
Therefore, by monotonicity of the infimum,
\begin{equation}
	\inf_{(w,\theta)\in\mathcal{W}\times\Theta} \widetilde{\mathcal{R}}(w,\theta)
	\;\le\;
	\inf_{w\in\mathcal{W}}\mathcal{R}(w,\theta_0).
	\label{eq:infimum}
\end{equation}
This inequality formalizes the fact that optimizing intrinsic neuron parameters cannot
worsen the best achievable expected risk.

\paragraph{Neuron-model-dependent descent directions}
To explain why improvements are typically observed in practice, we analyze how intrinsic
parameters enter the neuron dynamics and interact with learning algorithms.

For a discrete-time leaky integrate-and-fire neuron, the membrane potential evolves as
\begin{equation}
	V_m(t_k) = \left(1 - \frac{\Delta t}{\tau_m}\right) V_m(t_{k-1})
	+ w^\top x(t_{k-1}) - V_{th}
\end{equation}

where $\quad t_k = k \Delta$.

Thus, intrinsic parameters $\theta=(\tau_m,V_{\mathrm{th}})$ enter the dynamics affinely:
$V_{\mathrm{th}}$ acts as an additive bias, while $\tau_m$ controls the effective temporal
integration window.

Let $\widetilde{\mathcal R}(w,\theta)$ denote the surrogate training objective minimized by
the learning algorithm (e.g.\ backpropagation with surrogate gradients, Tempotron learning,
or gradient-based approximations of STDP).
By the chain rule,
\begin{equation}\label{7}
	\frac{\partial 	\widetilde{\mathcal{R}}(w,\theta)
	}{\partial V_{th}}
	= \sum_{k=0}^{T} \frac{\partial 	\widetilde{\mathcal{R}}(w,\theta)
	}{\partial V_m(t_k)} \cdot (-1)
	= - \sum_{k=0}^{T} \frac{\partial 	\widetilde{\mathcal{R}}(w,\theta)
	}{\partial V_m(t_k)}
\end{equation}

\begin{equation}\label{8}
	\frac{\partial 	\widetilde{\mathcal{R}}(w,\theta)
	}{\partial \tau_m}
	= \sum_{k=0}^{T}
	\frac{\partial 	\widetilde{\mathcal{R}}(w,\theta)
		\widetilde{\mathcal{R}}(w,\theta)
	}{\partial V_m(t_k)}
	\frac{\Delta t}{\tau_m^2} V_m(t_{k-1})
\end{equation}

Hence, unless the baseline configuration $\theta_0$ is already stationary, the gradient
with respect to intrinsic parameters is non-zero.

\paragraph{Algorithmic implication}
Under standard smoothness assumptions on $\widetilde{\mathcal R}(w,\cdot)$, a projected
gradient update of the intrinsic parameters,
\begin{equation}
	\theta^{+}
	=
	\Pi_{\Theta}\!\big(
	\theta_0-\eta\nabla_\theta\widetilde{\mathcal R}(w,\theta_0)
	\big),
	\qquad
	0<\eta\le \frac{1}{L_\theta},
\end{equation}
satisfies the descent inequality
\begin{equation}
	\widetilde{\mathcal R}(w,\theta^{+})
	\le
	\widetilde{\mathcal R}(w,\theta_0)
	-
	\frac{\eta}{2}
	\|\nabla_\theta\widetilde{\mathcal R}(w,\theta_0)\|^2.
\end{equation}

\paragraph{Role of $L_\theta$}
The constant $L_\theta$ denotes the Lipschitz constant of the gradient of the surrogate
training objective with respect to the intrinsic neuron parameters $\theta$.
Formally, it is defined by the smoothness condition

\begin{equation}
	\left\|
	\nabla_{\theta} \widetilde{\mathcal{R}}(w,\theta)
	(\mathbf{w}, \theta_1)
	-
	\nabla_{\theta} \widetilde{\mathcal{R}}(w,\theta)
	(\mathbf{w}, \theta_2)
	\right\|
	\le
	L_\theta
	\left\|
	\theta_1 - \theta_2
	\right\|,
\end{equation}

which ensures that the gradient of $\widetilde{\mathcal{R}}(w,\theta)
$ does not change arbitrarily fast in the directions corresponding to intrinsic parameters.

This assumption plays a central role in establishing the descent inequality for
gradient-based updates of $\theta$. Under the above smoothness condition, the standard
quadratic upper bound for $	\widetilde{\mathcal{R}}(w,\theta)
$ yields

\begin{equation}
	\widetilde{\mathcal{R}}(w,\theta)
	(\mathbf{w}, \theta^{+})
	\le
	\widetilde{\mathcal{R}}(w,\theta)
	(\mathbf{w}, \theta)
	\\
	-
	\left(
	\eta - \frac{L_\theta \eta^2}{2}
	\right)
	\left\|
	\nabla_{\theta} 	\widetilde{\mathcal{R}}(w,\theta)
	(\mathbf{w}, \theta)
	\right\|^2,
\end{equation}

where
$\quad \theta^{+} = \theta  - \eta \nabla_{\theta} \widetilde{\mathcal{R}}(w,\theta) (\mathbf{w}, \theta)$.

Consequently, choosing a learning rate
$\eta \le 1 / L_\theta$
guarantees a monotonic decrease of the surrogate objective.

Quantitatively, for the maximal admissible step size
$\eta = 1 / L_\theta$,
the decrease in the surrogate loss per update is bounded from below as

\begin{equation}\label{13}
	\widetilde{\mathcal{R}}(w,\theta)
	(\mathbf{w}, \theta)
	-
	\widetilde{\mathcal{R}}(w,\theta)
	(\mathbf{w}, \theta^{+})
	\ge
	\frac{1}{2 L_\theta}
	\left\|
	\nabla_{\theta} \widetilde{\mathcal{R}}(w,\theta)
	(\mathbf{w}, \theta)
	\right\|^2.
\end{equation}

Thus, $L_\theta$ directly controls how changes in intrinsic parameters translate
into guaranteed improvements of the training objective: smaller values of $L_\theta$
allow for larger learning rates and stronger descent guarantees, while larger values
indicate increased curvature or sensitivity of the objective with respect to intrinsic
neuron dynamics.

In the context of spiking neural networks, the magnitude of $L_\theta$ is influenced
by the temporal depth of backpropagation through time, the smoothness of the surrogate
gradients used to approximate spike discontinuities, and the stability properties of
the underlying neuron model. For this reason, $L_\theta$ is treated as a technical
smoothness constant rather than an explicitly computed quantity, serving to formalize
the conditions under which learning intrinsic neuron parameters yields a guaranteed
descent of the surrogate objective.

Consequently, learning intrinsic neuron parameters yields a strict decrease of the training objective whenever the baseline configuration is not stationary.

\paragraph{Extension to the meta-neuron model}
In the proposed meta-neuron architecture, each unit consists of multiple parallel LIF
sub-neurons with independent intrinsic parameters whose outputs are aggregated.
The standard LIF neuron corresponds to a special case of the meta-neuron with a single
sub-unit. Therefore, the set of feasible intrinsic-parameter descent directions in the
meta-neuron model strictly contains that of the LIF model, providing additional degrees of
freedom for optimization.

Combining the static inequality~\eqref{eq:infimum} with the algorithmic descent property
shows that optimizing intrinsic neuron parameters not only cannot worsen the achievable
risk, but generically improves it under the learning rules considered.
This provides a theoretical explanation for the substantial accuracy gains observed in
Tables~\ref{tab:accuracy_1st_db} and \ref{tab:accuracy_2st_db}, while preserving the same
asymptotic inference-time complexity.

\paragraph{Role of Lempel-Ziv complexity}

The risk inequality (\ref{13}) follows from hypothesis class inclusion and
does not depend on any specific properties of the complexity measure used in the decision
stage. Nevertheless, the choice of Lempel--Ziv complexity (LZC) plays a crucial and
structurally beneficial role in the proposed framework, as it determines how the additional
degrees of freedom introduced by intrinsic neuron parameters are translated into improved
classification performance.

Recall that the classifier has the form
\begin{equation}
	h_{w,\theta}(x)
	=
	g\!\Big(c_2\big(\Phi_{w,\theta}(x)\big)\Big),
\end{equation}
so that the entire spiking network dynamics is mapped onto a single scalar statistic
$c_2(\cdot)$ before the final decision is made.
As a consequence, any effect of synaptic weights $w$ and intrinsic parameters $\theta$ on
the classifier must act through their influence on the distribution of
$c_2(\Phi_{w,\theta}(X))$.

Intrinsic neuron parameters such as firing thresholds and decay constants directly affect
the temporal structure of spike trains generated by the network.
In particular, threshold adaptation modifies spike density and burst formation, while decay
constants control the effective temporal integration window and, hence, the degree of
temporal correlation between spikes.
Lempel-Ziv complexity is specifically sensitive to these properties, as it quantifies the
rate at which new substrings appear during sequential parsing of a binary sequence.
For sufficiently long sequences, $c_2(\cdot)$ provides an estimator closely related to the
entropy rate, while remaining sensitive to deterministic temporal regularities and
repetition patterns.

From a statistical perspective, the use of LZC induces a mapping
\begin{equation}
	\Phi_{w,\theta}(X)
	\;\longmapsto\;
	c_2(\Phi_{w,\theta}(X)) \in \mathbb{R},
\end{equation}
which reduces the high-dimensional spatiotemporal spike output of the network to a
one-dimensional summary statistic.
Learning intrinsic neuron parameters thus acts to reshape the class-conditional
distributions
\begin{equation}
	c_2(\Phi_{w,\theta}(X)\mid Y=0)
	\qquad\text{and}\qquad
	c_2(\Phi_{w,\theta}(X)\mid Y=1),
\end{equation}
by systematically increasing or decreasing temporal regularity in a class-dependent
manner.
Because separability in one dimension is determined entirely by the relative position and
spread of these distributions, even moderate changes in intrinsic parameters can lead to
substantial gains in classification accuracy.

Importantly, this increase in representational flexibility does not entail an increase in
decision complexity.
While optimizing $\theta$ enlarges the family of realizable spike transformations
$\Phi_{w,\theta}$, the computational cost of the decision stage remains unchanged, as the
evaluation of $c_2(\cdot)$ scales linearly with sequence length and is independent of the
number of intrinsic parameters.
This explains why the extended configurations achieve significantly higher accuracy without
altering the asymptotic inference-time complexity, as observed empirically in
Tables~\ref{tab:accuracy_1st_db}--\ref{tab:efficiency_metrics}.

In summary, Lempel-Ziv complexity does not directly contribute to the risk inequality
itself, but it provides a low-dimensional, information-theoretically meaningful interface
through which neuron-intrinsic degrees of freedom are effectively exploited by the learning
algorithms.
This alignment between neuron dynamics, complexity-based encoding, and statistical
separability is a key factor underlying the empirical performance gains reported in this
study.

Consequently, the baseline configuration $\theta_0$ is stationary with respect to intrinsic parameters if and only if both sums (\ref{7}) and (\ref{8}) vanish simultaneously. 
Unless this non-generic condition is satisfied, the gradient with respect to intrinsic
parameters is non-zero, and a gradient-based update of $\theta$ yields a strict decrease of
the surrogate objective.

Learning intrinsic neuron parameters therefore induces systematic deformations of the
class-conditional distributions of the scalar random variable
$Z_\theta = c_2(\Phi_{w,\theta}(X))$.
For each class $y\in\{0,1\}$, let
$\mu_{y,\theta} = \mathcal{L}(Z_\theta \mid Y=y)$ denote the corresponding conditional law.
Adjusting intrinsic parameters such as firing thresholds and decay constants modifies the
temporal regularity, burst structure, and correlation patterns of spike trains, which in
turn leads to continuous, class-dependent changes in the mean, variance, and shape of
$\mu_{y,\theta}$.

Because classification is performed in a one-dimensional feature space, the achievable
separability of the two classes is entirely determined by the relative position and spread
of $\mu_{0,\theta}$ and $\mu_{1,\theta}$.
In particular, the Bayes risk depends only on the overlap of these distributions, and even
moderate shifts in their means or reductions in their variances can lead to a pronounced
decrease in misclassification probability.
This explains why comparatively small adjustments of intrinsic neuron parameters may result
in substantial gains in classification accuracy when using Lempel--Ziv complexity as the
decision statistic.

\section{Discussion}
\par
This study provides one of the first systematic comparisons of biologically inspired meta-neurons and classical LIF models in spiking neural networks, analyzing how their performance interacts with learning rules and input encoding strategies. By jointly exploring neuron dynamics, training algorithms, and Poisson-based input variability, our results offer novel insights into the trade-offs between expressiveness, stability, and accuracy in SNN classification tasks.

The LIF-based networks consistently achieve high accuracy, with extended configurations reaching up to 99.50\% for Tempotron and above 96.00\% for both backpropagation and STDP, confirming their robustness under well-aligned Poisson input encodings \cite{Roy2019,Neftci2019}. In contrast, the meta-neuron architecture attains accuracies of up to 97.50\% with backpropagation and 95.50\% with STDP, but exhibits a higher sensitivity to input stochasticity and hyperparameter choices, consistent with stability challenges discussed in \cite{Cheng2023}. Backpropagation improves meta-neuron performance from 89.00\% to 97.50\% (an increase of 8.50 percentage points), but does not close the gap to the best LIF-based Tempotron configuration. This aligns with the broader challenge of fully exploiting meta-neuron expressiveness without incurring overfitting or instability \cite{Bellec2020}.

Beyond the empirical findings, the observed performance differences between LIF- and meta-neuron-based SNNs can be interpreted through theoretical and biological considerations. Classical LIF neurons implement a stable low-pass temporal integration mechanism, effectively smoothing variability in the incoming Poisson spike trains \cite{Gerstner2014}. This makes them well suited for tasks where discriminative information is encoded in coarse firing-rate differences. Their fixed membrane time constant and threshold act as implicit regularizers that suppress sensitivity to noise and reduce the dimensionality of the encoding space. This property explains why the LIF-based networks reach very high accuracies (above 96-99.00\%) and remain robust across different learning rules.

In contrast, meta-neurons introduce additional degrees of freedom by allowing thresholds and decay constants to vary or by aggregating multiple LIF sub-units \cite{Cheng2023,Yao2024}. From a dynamical systems perspective, this increases the dimensionality and expressiveness of the neuron model, enabling richer temporal selectivity and potentially sharper discrimination of complex spike patterns, but also making the dynamics more sensitive to perturbations \cite{Gerstner2014}. However, it also destabilizes the dynamics: meta-neurons respond more strongly to stochastic fluctuations in spike timing, which is consistent with the well-known sensitivity of high-dimensional recurrent or nonlinear spiking models to noise. Biologically, similar trade-offs appear in cortical circuits, where neurons with adaptive thresholds or multiple synaptic integration zones exhibit both higher computational capability and increased susceptibility to variability.

The improvements observed when training thresholds and decay parameters corroborate another principle of biological learning: intrinsic plasticity. Experimental studies show that real neurons adapt not only synaptic weights, but also excitability, firing thresholds, and membrane time constants to stabilize activity and maximize information transmission. In our experiments, enabling such intrinsic plasticity leads to accuracy gains of 7-10 percentage points for both LIF and meta-neuron models, reflecting the biological role of excitability regulation in optimizing neural computations under variable input statistics.

The application of Lempel-Ziv Complexity as a downstream readout offers an additional theoretical perspective on the obtained results. LZC provides an asymptotically consistent estimator of the entropy rate for stationary ergodic sources, and thus quantifies the balance between regularity and randomness in the output spike trains \cite{Shannon1948,LempelZiv1976,Bossomaier2016}. In our experiments, high-performing configurations produce output patterns with intermediate LZC values: sequences that are neither trivially regular (low complexity) nor indistinguishable from noise (maximal complexity). This is consistent with the notion that effective neural codes in both biological and artificial systems operate near a regime of maximal usable variability, where spikes are sufficiently structured to carry class-specific information, yet exhibit enough diversity to avoid saturation and redundancy \cite{Pregowska2019}. Moreover, the fact that the LZC-based classifier can reach accuracies up to 99.50\% indicates that much of the discriminative information generated by the SNN is captured at the level of structural complexity alone, without relying on fine-grained spike timing features in the final decision stage. This supports the view that SNNs can be used as powerful, biologically motivated feature extractors that transform raw Poisson spike trains into representations whose complexity profiles separate classes in a low-dimensional information-theoretic space. In turn, this suggests that complexity-based measures such as LZC may serve as a compact and interpretable interface between high-dimensional spiking dynamics and downstream machine learning or neuromorphic classifiers.

From a systems neuroscience perspective, the results highlight the potential of spiking neural networks not only as engineering tools, but also as computational models for probing fundamental questions of neural coding and learning. The observed sensitivity of classification performance to input stochasticity and spike timing mirrors variability found in biological systems \cite{Gerstner2014}, suggesting that SNNs can provide explanatory value for understanding sensory processing, neural variability, and adaptive behavior in biological circuits.
\par
In practical terms, these insights can inform the design of neuromorphic architectures for low-power, real-time applications such as edge computing, autonomous systems, and brain–computer interfaces \cite{Roy2019}. The complementary characteristics of stable LIF dynamics and adaptable meta-neuron components point toward hybrid architectures that balance interpretability with task-specific adaptability, i.e., well-suited for dynamic environments and non-stationary input streams. Furthermore, the strong dependence of expressive neuron models on precise hyperparameter tuning highlights the need for automated optimization pipelines \cite{Rudnicka2026}. Approaches such as meta-learning and biologically inspired meta-plasticity \cite{Bellec2020} represent promising avenues to improve convergence, generalization, and adaptability without extensive manual calibration.

\section{Conclusion}
This study provides new insights on how the choice of learning algorithm, neuronal parameter optimization, and input encoding jointly shape the classification performance of spiking neural networks for temporally structured data. Although biologically inspired mechanisms such as STDP and the Tempotron rule offer valuable insights into synaptic plasticity, our results show for the first time that combining gradient-based methods such as backpropagation with flexible, parameter-optimized neuron models can yield significantly higher accuracy. Moreover, we also observed that under well-structured input datasets, simpler LIF-based architectures can outperform more complex models, revealing an important trade-off between biological plausibility, model complexity, and real-world performance.

\section*{Conflict of interest} 
The authors declare that they have no known competing financial interests or personal relationships that could have appeared to influence the work reported in this paper.

\section*{Data availability}
Not applicable.

\section*{Author contribution} 
All authors contributed to the conception and design of the study. All authors performed material preparation, data collection, and analysis. The first draft of the manuscript was written by all authors who commented on previous versions of the manuscript. All authors read and approved the final manuscript.


\begin{thebibliography}{38}
	
	\bibitem{Tang2025}
	X. Tang, T. Chen, Q. Cheng, H. Shen, S. Duan, L. Wang, ``Spatio-temporal channel attention and membrane potential modulation for efficient spiking neural network, ''\textit{Engineering Applications of Artificial Intelligence}, vol. 148, 110131, 2025. \url{https://doi.org/10.1016/j.engappai.2025.110131}.
	
	\bibitem{Stan2024}
	M. I. Stan, O. Rhodes, `` Learning long sequences in spiking neural networks, '' \textit{Scientific Reports}, vol. 14, 21957, 2024. \url{https://doi.org/10.1038/s41598-024-71678-8}.
	
	\bibitem{Gao2025}
	R. Gao, C. Jiang, Y Zhang, ``Recurrent spiking neural networks as models of the entorhinal-hippocampal system for path integration: Grid cells and beyond,'' \textit{Neurocomputing} 651, C, 2025. \url{https://doi.org/10.1016/j.neucom.2025.130814}.
	
	\bibitem{Wu2018}
	Y. Wu, L. Deng, G. Li, J. Zhu, L. Shi, ``Spatio-temporal backpropagation for training high-performance spiking neural networks,'' \textit{Frontiers in Neuroscience} 12, 331, 2018. \url{https://doi.org/10.3389/fnins.2018.00331}.
	
	\bibitem{Yi2023}
	Z. Yi, J. Lian, Q. Liu, H. Zhu, D. Liang, J. Liu, ``Learning rules in spiking neural networks: A survey,'' \textit{Neurocomputing}, 531, 163--179, 2023. \url{https://doi.org/10.1016/j.neucom.2023.02.026}.
	
	\bibitem{Hu2025}
	Y. Hu, L. Deng, Y. Wu, M. Yao, G. Li, G,  ``Advancing spiking neural networks toward deep residual learning,'' \textit{IEEE Transactions on Neural Networks and Learning Systems} 36(2),2353--2367,2025. \url{https://doi.org/10.1109/TNNLS.2024.3355393}.
	
	\bibitem{bassler2022}
	D. Bassler, T. Kortus, G. Guhring, G. ``Unsupervised anomaly detection in multivariate time series with online evolving spiking neural networks,'' \textit{ Machine Learning} 111, 1377--1408, 2022. \url{https://doi.org/10.1007/s10994-022-06129-4}.
	
	\bibitem{Ding2025}
	J. Ding, J. Zhang, T. Huang, J. K. Liu, Z. Yu, ``Assisting training of deep spiking neural networks with parameter initialization, ''\textit{IEEE Transactions on Neural Networks and Learning Systems}, 2025. \url{https://doi.org/10.1109/TNNLS.2025.3547774}.
	
	\bibitem{Saglam2023}
	B. Saglam, S. S. Kozat, S.S., ``Deep intrinsically motivated exploration in continuous control,'' \textit{Machine Learning} 112, 4959--4993, 2023. \url{https://doi.org/10.1007/s10994-023-06363-4}.
	
	\bibitem{Rudnicka2024}
	Z. Rudnicka, J. Szczepanski, A. Pregowska, ``Artificial intelligence-based algorithms in medical image scan segmentation and intelligent visual content generation-A concise overview,'' \textit{Electronics} 13, 746, 2024. \url{https://doi.org/10.3390/electronics13040746}.
	
	\bibitem{Jin2025}
	H. Jin, X. Yang, S. Song, Z. Song, J. Ji, ``A temporally coded multilayer spiking neural network and its memristor-based hardware implementation,'' \textit{Neurocomputing} 656, 131523, 2025. \url{https://doi.org/10.1016/j.neucom.2025.131523}.
	
	\bibitem{Shannon1948}
	C. E. Shannon, ``A mathematical theory of communication,'' \textit{Bell System Technical Journal} 2,379--423,623–-656, 1948. \url{https://doi.org/10.1002/j.1538-7305.1948.tb01338.x}.
	
	\bibitem{LempelZiv1976}
	J. Ziv, A. Lempel, ``On the complexity of finite sequences,'' \textit{IEEE Transactions on Information Theory} 22(1), 75--81, 1976. \url{https://doi.org/10.1109/TIT.1976.1055501}.
	
	\bibitem{Bossomaier2016}
	T. Bossomaier, L. Barnett, M. Harre, J. T. Lizier, An Introduction to Transfer Entropy. Springer, Cham, 2016. \url{https://doi.org/10.1007/978-3-319-43222-9}.
	
	\bibitem{Arnold2013}
	M.M. Arnold, J. Szczepanski, N. Montejo, J. N. Amigó, E. Wajnryb, M.V. Sanchez–Vives, ``Information content in cortical spike trains during brain state transitions,'' \textit{Journal of Sleep Research} 22, 13--21, 2013. \url{https://doi.org/10.1109/TIT.1976.1055501}.
	
	\bibitem{Pregowska2019}
	A. Pregowska, K. Proniewska, P. van Dam, J. Szczepanski, ``Using Lempel-Ziv complexity as effective classification tool of the sleep-related breathing disorders,'' \textit{Computer Methods and Programs in Biomedicine} 182, 105052, 2019. \url{https://doi.org/10.1016/j.cmpb.2019.105052}.
	
	\bibitem{Dutta2017}
	S. Dutta, V. Kumar, A. Shukla, et al., ``Leaky integrate and fire neuron by charge-discharge dynamics in floating-body MOSFET,'' \textit{Scientific Reports} 7, 8257, 2017. \url{https://doi.org/10.1038/s41598-017-07418-y}.
	
	\bibitem{Cheng2023}
	X. Cheng, T. Zhang, S. Jia, B. Xu, ``Meta neurons improve spiking neural networks for efficient spatio-temporal learning,'' \textit{Neurocomputing} 531,217--225, 2--23, 2023. \url{ https://doi.org/10.1016/j.neucom.2023.02.029}.
	
	\bibitem{Bansal2024}
	Y. Bansal, D. Lillis, M.T. Kechadi, ``A neural meta model for predicting winter wheat crop yield, '' \textit{Machine Learning} 113, 3771--3788, 2024. \url{https://doi.org/10.1007/s10994-023-06455}.
	
	\bibitem{Hao2020}
	Y. Hao, X. Huang, M. Dong, B. Xu, ``A biologically plausible supervised learning method for spiking neural networks using the symmetric STDP rule,'' \textit{Neural Networks} 121, 387--395, 2020. \url{https://doi.org/10.1016/j.neunet.2019.09.007}.
	
	\bibitem{Gutig2006}
	R. Gutig, H. Sompolinsky, ``The tempotron: A neuron that learns spike timing-based decisions,'' \textit{Nature Neuroscience} 9(3), 420--428, 2006. \url{https://doi.org/10.1038/nn1643}.
	
	\bibitem{Yaya2025}
	S. Y. A. Yarga, S. U. N. Wood, ``Accelerating spiking neural networks with parallelizable leaky integrate-and-fire neurons, '' \textit{Neuromorphic Computing and Engineering} 5(1), 014012, 2025. \url{https://doi.org/10.1088/2634-4386/adb7fe}.
	
	\bibitem{Yao2024}
	M. Yao, J. Hu, T. Hu, Y. Xu, Z. Zhou, Y. Tian, B.  Xu, G. Li, ``Spike-driven transformer v2: Meta spiking neural network architecture inspiring the design of next-generation neuromorphic chips, '' 2024. \url{arXiv preprint arXiv:2404.03663}.
	
	\bibitem{Kara2000}
	P. Kara, P. Reinagel, R. C. Reid, ``Low response variability in simultaneously recorded retinal, thalamic, and cortical neurons, '' \textit{Neuron} 27(3), 635--646, 2000. \url{https://doi.org/10.1016/s0896-6273(00)00072-6}.
	
	\bibitem{Barbieri2001}
	R. Barbieri, M. C. Quirk, L. M. Frank, M. A. Wilson, E. N. Brown, E. N., ``Construction and analysis of non-Poisson stimulus-response models of neural spiking activity,'' \textit{Journal of Neuroscience Methods} 105, 25--37, 2001. \url{https://doi.org/10.1016/s0165-0270(00)00344-7}. 
	
	\bibitem{Averbeck2009}
	B.B. Averbeck, ``Poisson or not Poisson: Differences in spike train statistics between parietal cortical areas,''N\textit{Neuron} 62(3), 310--311, 2009. \url{https://doi.org/10.1016/j.neuron.2009.04.021}.
	
	\bibitem{Maric2025}
	N.Maric, ``Stochastic model of seed dispersal with homogeneous and non-homogeneous Poisson processes under habitat reduction conditions, ''\textit{Journal of Biological Physics} 51, 1,2025.  \url{https://doi.org/10.1007/s10867-024-09666-2}.
	
	\bibitem{Singh2022}
	A. Singh, S. Kushwaha, M. Alarfaj, M. Singh,  ``Comprehensive overview of backpropagation algorithm for digital image denoising,'' \textit{Electronics} 11, 1590, 2022. \url{https://doi.org/10.3390/electronics11101590}.
	
	\bibitem{Kaur2022}
	B. Kaur, S. Kumar, B. Kumar Kaushik, ``Recent advancements in optical biosensors for cancer detection,'' \textit{Biosensors and Bioelectronics} 197, 113805, 2022. \url{https://doi.org/10.1016/j.bios.2021.113805}.
	
	
	\bibitem{Shah2019}
	D. Shah, G. Y. Wang, M. Doborjeh, Z. Doborjeh, N. Kasabov, ``Deep learning of EEG data in the NeuCube brain-inspired spiking neural network architecture for a better understanding of depression,'' in \textit{Neural Information Processing: Proceedings of ICONIP 2019}, Part III, 195--206, 2019. \url{https://doi.org/10.1007/978-3-030-36718-3\_17.}
	
	\bibitem{Markram2011}
	H. Markram, W. Gerstner, P.J. Sjöström, ``A history of spike-timing-dependent plasticity,'' \textit{Frontiers in Synaptic Neuroscience} 3, 4, 2011. \url{https://doi.org/10.3389/fnsyn.2011.00004}.
	
	\bibitem{Merolla2014}
	P. A. Merolla, J. V. Arthur, R. Alvarez-Icaza, A. S. Cassidy, J. Sawada, F. Akopyan, ``A million spiking-neuron integrated circuit with a scalable communication network and interface,'' \textit{Science} 345, 668--673, 2014. \url{https://doi.org/10.1126/science.1254642}.
	
	\bibitem{Lagani2023}
	G. Lagani, F. Falchi, C. Gennaro, G. Amato, ``Spiking neural networks and bio-inspired supervised deep learning: A survey,'' 2023. \textit{arXiv preprint arXiv:2307.16235}.
	
	\bibitem{Roy2019}
	K. Roy, A. Jaiswal, P. Panda, ``Towards spike-based machine intelligence with neuromorphic computing,'' \textit{Nature} 575, 607--617, 2019. \url{https://doi.org/10.1038/s41586-019-1677-2}.
	
	\bibitem{Rudnicka2026}
	Z. Rudnicka, J. Szczepanski, A. Pregowska, ``Impact of Neuron Models on Spiking Neural Network Performance: A Complexity-based Classification Approach,'' \textit{Neuroinformatics} 24, 5, 1--23, 2026. \url{https://doi.org/10.1007/s12021-025-09754-1}.
	
	\bibitem{Neftci2019}
	E. P. Neftci, H. Mostafa, F. Zenke, ``Surrogate Gradient Learning in Spiking Neural Networks: Bringing the Power of Gradient-Based Optimization to Spiking Neural Networks,'' \textit{IEEE Signal Processing Magazine} 36(6), 51--63, 2019. \url{https://doi.org/10.1109/MSP.2019.2931595}.
	
	\bibitem{Bellec2020}
	G. Bellec, F. Scherr, A. Subramoney, A. et al. (2020). ``A solution to the learning dilemma for recurrent networks of spiking neurons.,'' \textit{Nat Commun} 11, 3625, 2020. \url{https://doi.org/10.1038/s41467-020-17236-y}.
	
	\bibitem{Gerstner2014}
	W. Gerstner, W. et al., Neuronal Dynamics: From Single Neurons to Networks and Models of Cognition. Cambridge University Press. \url{https://doi.org/10.1017/CBO9781107447615}.
\end{thebibliography}
\end{document}